\documentclass[preprint,12pt]{elsarticle}

\makeatletter
\def\ps@pprintTitle{%
\let\@oddhead\@empty
\let\@evenhead\@empty
\def\@oddfoot{}%
\let\@evenfoot\@oddfoot}
\makeatother

\usepackage{graphicx}
\usepackage{tikz}
\usepackage{comment}
\usepackage{amsmath,amssymb} 
\usepackage{color}
\usepackage[accsupp]{axessibility}
\usepackage{soul}
\usepackage[super]{nth}
\usepackage{multirow}
\usetikzlibrary{calc}
\usepackage[export]{adjustbox}
\usepackage{gensymb}
\usepackage{pgfplots}
\usepackage{picins}

\pgfplotsset{width=6.0cm,height=4.1cm,compat=1.9}

\newcommand{\multi}[1]{\textcolor{black}{#1}}

\renewcommand{\thefootnote}{\alph{footnote}}
\newcommand{\astfootnote}[1]{%
\let\oldthefootnote=\thefootnote%
\setcounter{footnote}{0}%
\renewcommand{\thefootnote}{\fnsymbol{footnote}}%
\footnote{#1}%
\let\thefootnote=\oldthefootnote%
}

\begin{document}
	
	\title{Bottom-Up 2D Pose Estimation
	via Dual Anatomical Centers
	for Small-Scale Persons}
	
	\author[1]{Yu~Cheng\fnref{fn1}}
	\author[1]{Yihao~Ai\fnref{fn1}\corref{cor1}}
	\author[2]{Bo~Wang}
	\author[1]{Xinchao~Wang}
	\author[1]{Robby~T.~Tan}
	
	\fntext[fn1]{Co-author}
	\cortext[cor1]{Corresponding author. Email: e0724394@u.nus.edu}
	
	\address[1]{Electrical and Computer Engineering, National University of Singapore, Singapore}
	
	\address[2]{CtrsVision, USA}
	
		
	


\begin{abstract}
	%
	In multi-person 2D pose estimation, the bottom-up methods simultaneously predict poses for all persons, and unlike the top-down methods, do not rely on human detection. 
	However, the SOTA bottom-up methods' accuracy is still inferior compared to the existing top-down methods. 
	This is due to the predicted human poses being regressed based on the inconsistent human bounding box center and the lack of human-scale normalization, leading to the predicted human poses being inaccurate and small-scale persons being missed. 
	To push the envelope of the bottom-up pose estimation, we firstly propose multi-scale training to enhance the network to handle scale variation with single-scale testing, particularly for small-scale persons. 
	Secondly, we introduce dual anatomical centers (i.e., head and body), where we can predict the human poses more accurately and reliably, especially for small-scale persons.
	Moreover, existing bottom-up methods use multi-scale testing to boost the accuracy of pose estimation at the price of multiple additional forward passes, which weakens the efficiency of bottom-up methods, the core strength compared to top-down methods. By contrast, our multi-scale training enables the model to predict high-quality poses in a single forward pass (i.e., single-scale testing).
	Our method achieves 38.4\% improvement on bounding box precision and 39.1\% improvement on bounding box recall over the state of the art (SOTA) on the challenging small-scale persons subset of COCO. For the human pose AP evaluation, we achieve a new SOTA (71.0 AP) on the COCO test-dev set with the single-scale testing. We also achieve the top performance (40.3 AP) on the OCHuman dataset in cross-dataset evaluation.

\end{abstract}
\begin{keyword}
	Multi-person pose estimation, human pose estimation, anatomical centers
\end{keyword}

\maketitle

\begin{highlights}
\item Research highlight 1
\item Research highlight 2
\end{highlights}

\section{Introduction}
\label{sec:intro}

Human 2D pose estimation is a fundamental research topic in computer vision and has a broad impact on a great number of applications such as human action understanding, person image generation, augmented reality, and motion capture \cite{yan2018spatial,ma2017pose,weng2019photo,dong2019towards,desmarais2021review}. 
Significant progress has been made by employing deep learning in recent years \cite{toshev2014deeppose,pfister2015flowing,newell2016stacked,pishchulin2016deepcut,insafutdinov2016deepercut,cao2017realtime,newell2017associative,cao2019openpose,sun2019deep}, yet multi-person 2D pose estimation from a single image is still an open problem.

There are two paradigms addressing the problem: top-down and bottom-up.
The major difference between the two is the use of human detection and the follow-up sequential processes. 
The top-down methods \cite{he2017mask,chen2018cascaded,xiao2018simple,tian2019directpose,sun2019deep} employ human detection to extract and crop each person into individual image patches, and then apply pose estimation to each patch sequentially.
As each image patch is normalized, the scale variation of each person is largely reduced, which is suitable for convolutional neural network (CNN) training. 
While with the help of human detection, the top-down methods enjoy higher accuracy than the bottom-up methods for human pose estimation, 
but it is still an open problem for human detection in multi-person settings.
Moreover, existing top-down methods process each person's image patch sequentially, which makes them inapplicable and unscalable in crowded scenarios.

The bottom-up methods \cite{cao2019openpose,newell2017associative,kocabas2018multiposenet,papandreou2018personlab,cheng2020higherhrnet,DEKR2021,luo2021rethinking,braso2021center,wang2021robust} process the whole image at once, which simultaneously estimate all possible keypoints belonging to every person, and then group the keypoints together to form individual persons' skeletons.
In the bottom-up paradigm, one forward pass is able to obtain all possible keypoints from an input image, and the follow-up grouping step associates the keypoints into individual skeletons, which is more efficient than the sequential patch-by-patch process of the top-down methods. 

\input{figure_tex/teaser_example}

However, since the input image is processed in one forward pass and no person-wise image normalization is utilized, the bottom-up multi-person pose estimation methods fail to handle human scale variation, leading to missing or wrongly predicted poses for small-scale persons, which is a fundamental problem in bottom-up multi-person pose estimation \cite{cheng2020higherhrnet}. Image pyramid for heatmap prediction \cite{papandreou2018personlab} and scale-aware representation using high-resolution feature pyramid \cite{cheng2020higherhrnet} have been developed to mitigate the problem.
Unfortunately, the existing bottom-up methods still constantly miss small-scale persons. Missing small-scale persons is a common problem in the SOTA bottom-up methods as the network lacks the capability of handling scale variation. Multi-scale testing is used in the SOTA bottom-up methods \cite{cheng2020higherhrnet,DEKR2021} to mitigate the problem, where multiple additional forward passes are needed, which clearly slows down inference. 

\begin{figure}
	\centering
	\includegraphics[width=\linewidth]{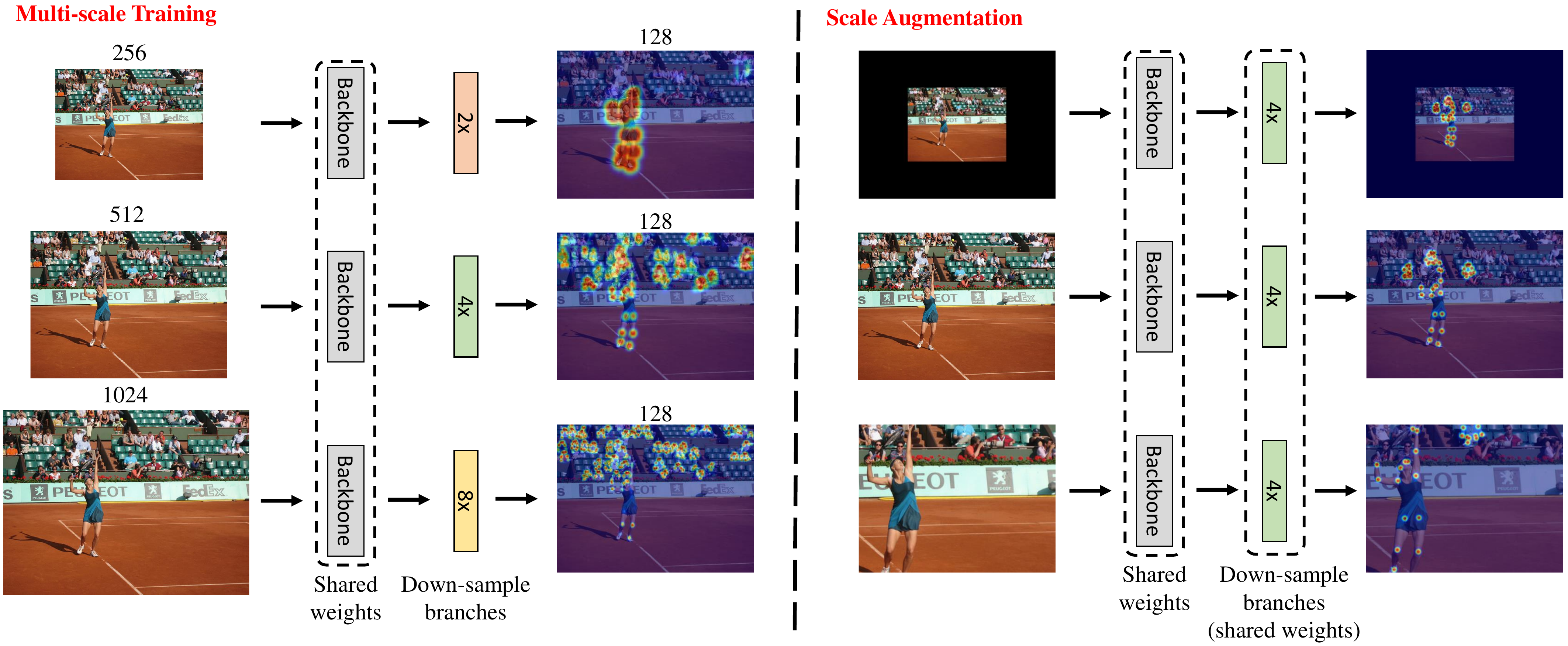}
	\caption{Difference between scale augmentation and our proposed \multi{multi-scale training.} Our multi-scale training utilizes different kernel sizes and multiple branches where each of them is responsible for a certain scale of persons, which makes it more adaptive and can better handle the scale variation compared to scale augmentation.}
	\label{fig:multi_scale_training}
\end{figure}

To solve the problem, we propose to use \multi{multi-scale training} to enhance the network's capability in handling persons across different scales. This multi-scale training allows the model to generate a more robust estimation in a single forward pass.
It also yields two advantages: (1) Since the annotations in training data contain limited scales of persons (mainly including medium and large scales), our \multi{multi-scale} training will broaden the scale variation of existing annotations and enhances the robustness against scale changes, especially the small-scale persons. (2) The \multi{multi-scale} property of network structure is specifically utilized to handle scale variation. 
As shown in Fig.\ref{fig:multi_scale_training}, our multi-scale training utilizing different kernel sizes for different human scales is more capable of dealing with the scale variation than the simple scale augmentation approach, where only one heatmap with a fixed kernel size is used. Moreover, our multi-scale training can better handle the scale variation than the existing methods that use only a single scale of heatmap \cite{luo2021rethinking} or a fixed kernel size \cite{cheng2020higherhrnet}.  

%

Besides solving the scale variation, our multi-scale training significantly improves the pose estimation for small-scale persons. 
However, the predicted poses of small-scale persons are more sensitive to the same deviation from the ground truth than that of large or medium-scale persons. 
Hence, to further improve the pose estimation for small-scale persons, it is important to ensure the deviation from the ground truths of the predicted poses is small, which is the problem of the SOTA bottom-up methods \cite{DEKR2021,wang2021robust}. 
The unreliable human center(s) (bounding box center) used to detect each person in these methods cause their predicted poses to be inaccurate, particularly for small-scale persons.
Specifically, because of the human center(s) (i.e., single \cite{DEKR2021} or multiple \cite{wang2021robust}) that are not defined anatomically are relatively arbitrary, the subsequent regression of the offsets of each human keypoint to the center(s) becomes inaccurate, resulting in error-prone human pose estimation.

To solve the aforementioned problem, we propose to use human anatomical centers to represent a person instead of the bounding box center(s) used in the existing methods \cite{DEKR2021,wang2021robust} for improving pose estimation accuracy. 
In particular, we found that directly regressing keypoints' location as offsets to the human bounding box center is not accurate because the location of the bounding box center is relatively arbitrary (i.e., not defined based on human anatomy). 
Therefore, we propose to use two anatomical landmarks (i.e., head and body centers) as human centers instead of the arbitrary center (i.e. bounding box center). 
Based on the two anatomical centers, we estimate two human poses for each person to increase the chance to capture a person, and then merge the two poses into one based on pose similarity that considers both spatial configuration and image appearance similarity.

Fig.~{\ref{fig:teaser_example}} shows a representative image with large scale variation, and the multi-person pose estimation results of our method, two SOTA bottom-up methods (DEKR {\cite{DEKR2021}} and SWAHR {\cite{luo2021rethinking}}), and the ground-truth human poses. This example clearly demonstrates that our method can correctly predict human poses of different scales. In contrast, the SOTA methods fail to predict (DEKR) or produce wrong human poses (SWAHR) for small-scale persons. More qualitative results are provided in Fig.~{\ref{fig:qualitative_result}} and \ref{fig:qualitative_whole_img} to illustrate that our method consistently outperforms the SOTA methods in multi-person pose estimation, particularly for small-scale persons. Our major contributions are summarized in the following:
	
	\begin{itemize}
		\item We propose to estimate dual anatomical centers, and then fuse the obtained poses together by pose confidence and similarity, replacing the unreliable human bounding box center estimation. 
		\item We introduce a multi-scale training framework, which handles persons across different scales, especially for small-scale persons. 
		\item We achieve a new SOTA 71.0 AP on the COCO test-dev set with single-scale testing, and \textbf{38.4\%} and \textbf{39.1\%} improvements on bounding box precision and recall respectively over the state-of-the-art methods on the challenging small-scale persons subset of COCO validation set.
	\end{itemize}
	
	\section{Related Work}
	\label{sec:related_work}
	
	Multi-person pose estimation has been developed rapidly with the help of deep learning in recent years. Earlier works propose to use CNN for human pose estimation from image \cite{tompson2014joint,toshev2014deeppose} or video \cite{pfister2015flowing,belagiannis2017recurrent,xiao2018simple}. Several popular network structures are developed including stacked hourglass networks \cite{newell2016stacked}, OpenPose \cite{wei2016convolutional,cao2017realtime,cao2019openpose}, and HRNet \cite{sun2019deep}. Different keypoints grouping approaches are developed for multi-person pose estimation such as DeepCut \cite{pishchulin2016deepcut,insafutdinov2016deepercut}, associative embedding \cite{newell2017associative}, part affinity fields \cite{cao2017realtime}, and PifPaf \cite{Kreiss_2019_CVPR}. 
	
	\vspace{0.3cm}
	\noindent \textbf{Top-Down Methods}  With the fast development in object detection \cite{girshick2014rich,girshick2015fast,ren2015faster,he2017mask}, top-down human pose estimation methods are proposed, where human detection is performed first to crop and normalize image patch for each person, and then pose estimation is applied to each image patch sequentially to predict the human pose \cite{he2017mask,chen2018cascaded,xiao2018simple,tian2019directpose,sun2019deep}. Top-down methods achieve high accuracy, which benefits from human detection and image normalization. As a result, each image patch contains one target person, and the scale of the person is normalized, which is suitable for network training. However, these methods require human detection and sequential person-by-person pose estimation, resulting in slow inference speed for images with multiple persons. In fact, the inference time is linear to the number of persons in an image.
	
	\vspace{0.3cm}
	\noindent \textbf{Bottom-Up Methods} Bottom-up methods simultaneously detect all keypoints in an image without using human detection \cite{cao2019openpose,newell2017associative,kocabas2018multiposenet,cheng2020higherhrnet,DEKR2021}. Compared to top-down methods, bottom-up methods process the whole image at once (i.e., one forward pass), thus, it can achieve fast inference. However, the efficiency brought by one forward passing affects the accuracy as persons from large to small scale are processed together at the same resolution, which makes the network difficult to train and perform poorly on small-scale persons. 
	The bottom-up paradigm can be realized in a two-stage manner, first to detect all keypoints at once by taking the whole image as input, and then group the keypoints together to form individual human poses \cite{newell2017associative,wei2016convolutional,cao2017realtime,cao2019openpose,Kreiss_2019_CVPR}. 
	
	Recently, DEKR \cite{DEKR2021} developed a multi-head framework, one head for keypoints' heatmap estimation, the other for human bounding box center and offset to replace grouping. But it also relies on multi-scale testing to boost accuracy. 
	CGNet \cite{braso2021center} proposes an attention mechanism to cluster each keypoint to its corresponding center, which helps in occlusion cases but still cannot handle small-scale persons. PINet \cite{wang2021robust} proposes to divide the human bounding box into three parts and estimate each part separately, which makes it more robust to body-part occlusion.  
	These human bounding box center-based methods \cite{DEKR2021,wang2021robust} that use bounding box center-based offset (i.e., keypoints to the center) regression show promising results. However, the bounding box only covers the visible part of a person, leading the bounding box center to be located arbitrarily over a person. This arbitrary location is used as the human center in the existing methods \cite{DEKR2021,wang2021robust} to regress the offset between each keypoint and the center, which increases the difficulty of the network to reliably and accurately learn the offset, thus, hurting the accuracy of the pose estimation, regardless one bounding box center is used like DEKR \cite{DEKR2021} or multiple bounding box centers are used like PINet \cite{wang2021robust}. 
	
	\begin{figure*}[t!]
		\centering
		\includegraphics[width=\textwidth]{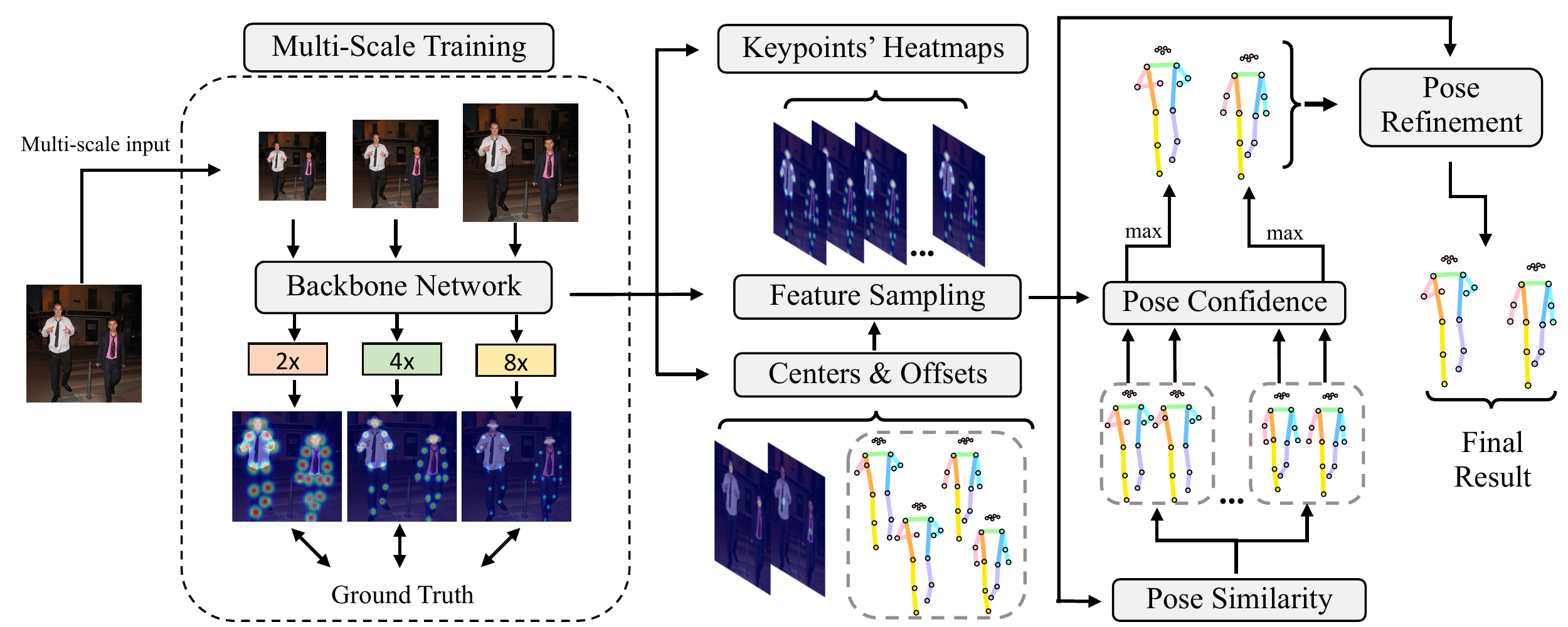}
		\caption{An overview of the proposed framework. Our method is composed of a backbone network with multi-scale training and two pose estimation branches, one for keypoints' heatmaps, the other for dual human centers and their offsets. Other major components include: 1) a pose similarity network that estimates pose similarity to group poses according to their identities. 2) a pose confidence network that predicts the confidence of a predicted pose, which is used to select the most confident pose out of a group of candidates. 3) a pose refinement network that merges the information from the predicted poses and multi-scale features for final human poses.}
		\label{fig:overview}
	\end{figure*}
	
	Without human detection and normalization like in top-down methods, handling human scale variation is a fundamental problem in the bottom-up methods. 
	Earlier works tackle the problem by proposing feature pyramid \cite{newell2017associative,papandreou2018personlab}, or skip connections \cite{nie2018pose}. 
	Recent work \cite{wei2020point} proposes to regress human poses based on pre-set anchors to deal with scale variation. HigherHRNet \cite{cheng2020higherhrnet} proposes an upsampling module to generate multi-scale heatmaps and aggregate them at inference time. SWAHR \cite{luo2021rethinking} introduces an adaptive Gaussian kernel to represent the scale of a person, which enables the model to be more robust to scale variance and labeling ambiguities. 
	The proposed solutions may help to enhance the pose estimation accuracy if small-scale persons are detected in the first place, however, the resolution for those persons is mostly insufficient, leading to many small-scale persons being missed. 
	
	\section{Proposed Method}
	\label{sec:method}
	
	Fig.~\ref{fig:overview} is an overview of the proposed framework. 
	Our backbone network takes \multi{multi-scale images} as input and produces \multi{multi-scale heatmaps.}
	Two major branches are built on top of the backbone network to handle the obtained \multi{multi-scale feature maps}: One branch is to estimate the keypoints' heatmaps, and the other branch is to estimate two human centers and their corresponding offsets (i.e., keypoint locations with respect to the centers). 
	%
	As we use two human centers for each person, resulting in two predicted human poses from the offsets for each person, the offset regression network may also generate false positives (i.e. duplicated human poses for the same person). Thus, the number of the obtained human poses usually is more than the number of persons in the image.
	
	To select a unique pose for each person, we propose a pose similarity network to make image appearance features from the same person close to each other, which is used to predict the similarity to determine if several poses belong to the same person.
	Furthermore, it is necessary to pick one out of several candidate poses. To this end, we introduce a pose confidence score that measures how close a predicted pose is to its corresponding ground truth. 
	With the pose confidence score, we can pick the pose with the highest score from each pose group. 
	Finally, the poses, the heatmaps, and the multi-scale features are fed into a pose refinement network to obtain the refined poses.
	
	\subsection{\multi{Multi-Scale Training}}
	\label{sec:mst}
	
	In bottom-up human pose estimation, as the whole image is processed with one forward pass to find all keypoints belonging to all possible persons, the resolution of small-scale persons is mostly insufficient to produce accurate pose estimation. 
	Existing bottom-up methods resort to the multi-scale testing to boost their accuracy \cite{cheng2020higherhrnet,DEKR2021}, which involves multiple additional forward passes of images with different resolutions at test time. 
	Thus, the improvement of the accuracy is at the cost of slowing the processing speed, where efficiency is one of the key advantages of the bottom-up methods over the top-down methods. 
	
	To address this problem, we propose to perform \multi{multi-scale training}, which utilizes multi-scale images as input for our network in the training stage. On the one hand, such an approach grants and enhances the model's robustness to the scale variation by broadening the scale diversity. In the multi-scale training process, the network takes the image with different sizes as input and obtains the heatmaps with different kernel sizes as output. This process makes each size of heatmap respond to different sizes of persons, i.e. large kernel size responds to larger persons and small kernel responds to smaller persons, which makes the model better capable of handling human-scale variation.
	The proposed multi-scale training takes a longer training time in exchange for better performance, where only one forward pass (i.e., single-scale testing) is needed at a time for desirable accuracy (e.g., 71.0 AP on the COCO test-dev is achieved, which is a new SOTA.). The general idea of the multi-scale training is illustrated in the left part of Fig.~\ref{fig:multi_scale_training}. Note HRNet backbone can take different sizes of input \cite{cheng2020higherhrnet,DEKR2021}.

	\vspace{0.3cm}
	\noindent \textbf{Two Branches of Pose Estimation} 
	As shown in Fig.~\ref{fig:overview}, in our framework, multi-person pose estimation is achieved using two branches.
	The first branch is to regress the heatmaps of each keypoint, which does not provide the identity information of the keypoints, and just estimates all possible keypoints for every person. 
	The second branch is to estimate human centers and the offset of each keypoint with respect to each center. 
	%
	The estimated offsets form individual poses $P$, but with duplication (i.e., partially due to the two human centers for each person, and partially due to the offset network's false positive estimates). 
	
	\subsection{Dual Anatomical Centers for Human Localization}
	\label{sec:two_centers}
	To obtain individual poses, existing bottom-up methods use either grouping (e.g., associative embedding \cite{newell2017associative}, Part Affinity Fields \cite{cao2019openpose}) or human center detection (e.g., human bounding box center and offset \cite{DEKR2021}).
	Compared with the grouping strategy, where the process is performed at a second stage after all keypoints are identified, estimating the human center and keypoints' offset to the center is more efficient,
	since the process is in parallel to the regression of the keypoints' heatmaps.
	However, we consider that in the context of multi-person pose estimation, 
	human anatomical keypoints as centers are more consistently defined in visual context across different scenarios than the human detection bounding box center used in the existing method \cite{DEKR2021}. 
	
    There are two benefits of using dual anatomical keypoints-based centers. 
	First, 
	the bounding box center is an arbitrary location depending on the visible part of the person \cite{DEKR2021}, 
	which is difficult for a network to learn.
	On the contrary, keypoints such as head or body center (i.e., the interpolation of shoulders and hips), have clear anatomical definitions, and thus their locations can be inferred based on the visual context.
	Hence, If a network is trained based on the dual anatomical centers, the network can learn more reliable and consistent centers. 
	Second, the anatomical centers-based offsets can be learned more reliably and accurately than the bounding box center-based ones. Due to the location of the bounding box centers' being relatively arbitrary, the corresponding offsets are more random, which is difficult to learn by the network. In contrast, the offsets of the anatomical centers can be learned more accurately due to human anatomy being well-defined, and therefore enabling the model to produce more accurate human pose prediction.

	Once the two centers for each person are detected, one person may have two centers and the same number of sets of corresponding offsets. 
	The network may also produce duplicated predictions, resulting in false positives in evaluation.
	To remove such duplication, we propose a pose similarity network to learn the appearance feature embedding of the predicted human pose.
	Given $M$ predicted poses $\{P^1, ..., P^M \}$ in an image, we assign their identity based on the closest ground truth, then generate positive and negative pose pairs (e.g., if two poses belong to the same person, they form a positive pose pair; otherwise, that's a negative pose pair). 
	The poses that are not close to any ground truth, are excluded from generating the pose pairs.
	
	In training, the network takes a pair of poses' corresponding image features and heatmaps as input, and outputs two pose feature vectors $f^i$ and $f^j$. 
	The cosine similarity of the two feature vectors is computed, and the network is trained with a binary label which is $1$ if they belong to the same person and $-1$ otherwise. 
	After the training, if two poses' features are similar to each other, the feature vectors should be close to each other.
	To compute the similarity of two human poses, we use a feature-based distance $D_{app} (f^i, f^j)$ from the pose similarity network, where $f^i$ is the embedded appearance feature of pose $P^i$, together with Gaussian distance of pose spatial configuration $D_{sp}(P^i, P^j)$, which is defined as:
	\begin{equation}
		D_{sp}(P^i, P^j) =  \frac{1}{N}\sum_{k=1}^N \frac{1}{\sigma\sqrt{2\pi}} 
		\exp\left( -\frac{1}{2}\left(\frac{P^i_k - P^j_k}{\sigma}\right)^{\!2}\,\right)
	\end{equation}
	where $P^i$ and $P^j$ are a pair of human poses, $P_k$ indicates the $k^{th}$ keypoint of the pose $P$, $N$ is the number of keypoints, $\sigma$ is the variance parameter of the Gaussian distance. 
	The pose similarity between a pair of two poses is computed using $D_{pose} \left( P^i, P^j \right) = D_{app}(f^i, f^j) D_{sp}(P^i, P^j)$. 
	With the proposed pose similarity $D_{pose}$, we can split the predicted poses into groups, and then pick the pose with the highest pose confidence, which is defined in the next section (Eq.~\ref{eq:pose_conf}). 
	
	
	\subsection{Pose Confidence Estimation}
	\label{sec:pose_confidence}
	
\begin{figure*}[t]
		\centering
		\includegraphics[width=\textwidth]{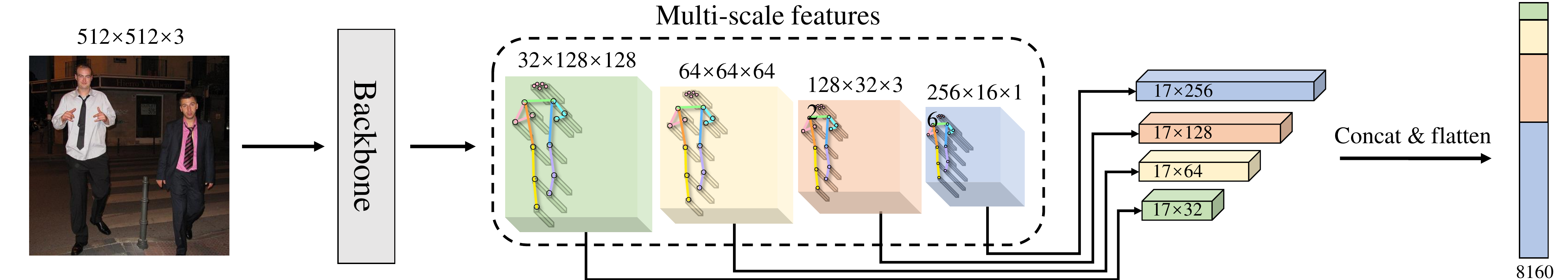}
		\caption{Illustration of how to obtain the sampled feature $F$. Black circles indicate the location to sample the feature maps across different scales based on the estimated human keypoints.}
		\label{fig:sampling}
	\end{figure*}

	Existing methods~\cite{DEKR2021,wang2021robust} use human bounding box center's confidence or the average of keypoints' confidence \cite{cheng2020higherhrnet} to measure the correctness of a predicted human pose in eliminating duplicate poses, which is not a reasonable choice because the confidence of a center/keypoint only reveals its visibility but not the likelihood of the predicted human pose. 
	To this end, we propose to regress the pose confidence, which effectively predicts the confidence of an estimated human pose. 
	We train a Multi-Layer Perceptron (MLP), that takes the concatenation of the sampled intermediate features $F$ (as illustrated in Fig.~\ref{fig:sampling}) and predicted heatmaps $H$ according to a predicted pose $P$ as input to regress the Object Keypoint Similarity (OKS) introduced in the COCO evaluation protocol \cite{lin2014microsoft} between ground-truth pose $\Tilde{P}$ and predicted pose $P$. 
	The loss of the pose confidence is defined as:
	\begin{equation}
		L_{pose} = | \text{OKS}(P,\Tilde{P}) - \text{MLP}(F \oplus H) |^2.
		\label{eq:pose_conf}
	\end{equation}
	
\begin{figure*}[t]
		\centering
		\includegraphics[width=\linewidth]{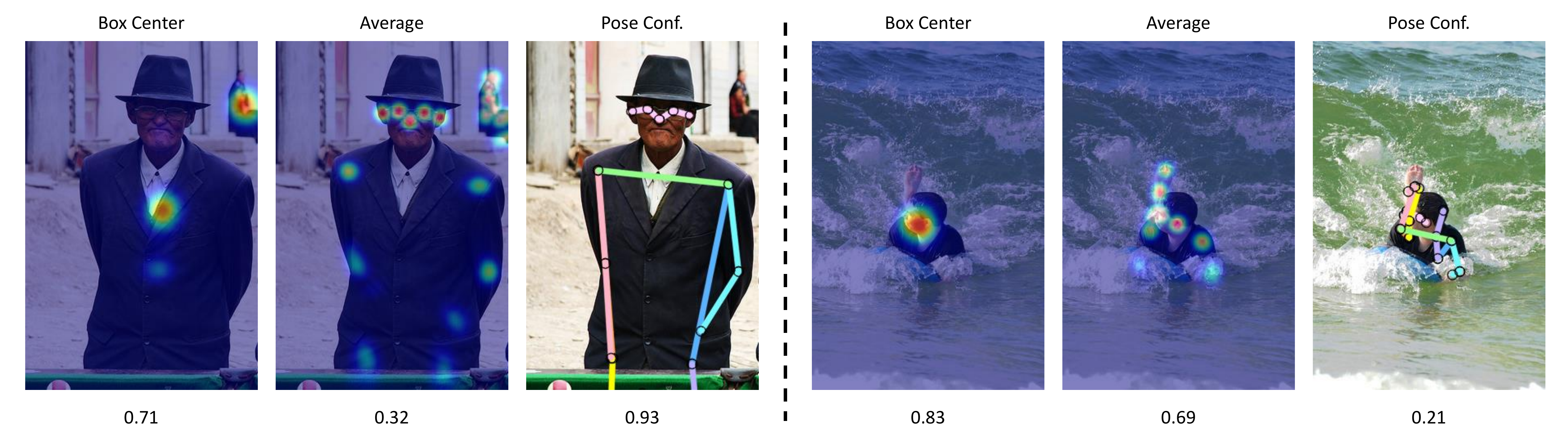}
		\caption{Comparisons of the proposed pose confidence with the existing approaches. Two examples are provided, the input image is shown as the background for each example. In each example, the heatmap of the bounding box center, the keypoints' heatmaps, and the predicted human pose are provided. The numbers under each image indicate the confidence value of each approach.}
		\label{fig:pose_confidence}
	\end{figure*}

	Note that for images with multiple persons, we choose the ground truth $\Tilde{P}$ closest to the predicted $P$ to calculate the OKS value. Two examples are provided in Fig.~\ref{fig:pose_confidence} to show the difference between the existing approaches (i.e., bounding box confidence \cite{DEKR2021} or average of keypoints' confidence \cite{cheng2020higherhrnet}) and our pose confidence. We observe that the confidence score of the bounding box center is irrelevant to the correctness of the predicted human pose as shown in the right example (i.e., the predicted pose is wrong), and sensitive to occlusion as shown in the left example (i.e., lower confidence value caused by occlusion). The average of each keypoint's confidence value as an alternative, is affected by occlusion (left) and rare poses (right) as well. 
	In contrast, our pose confidence reflects the correctness of the predicted human pose, at the same time, it is robust to occlusion.
	
	
 	\vspace{0.3cm}
	\noindent \textbf{Pose Refinement}
	Unlike most existing methods, we refine our pose estimation further.
	We use cross-scale information instead of simply merging the regression-based poses and the heatmap-based poses \cite{zhou2019objects}, where no additional information is available to fix the errors in the predicted human poses. 
	Based on our proposed pose similarity and pose confidence, we obtain the regression-based poses $\{{P}^1,... ,{P}^K\}$  from the estimated centers and offsets. 
	%
	For refinement, we estimate the residual between predicted poses and ground truths, using sampled features across scales to integrate multi-scale information. 
	We use the multi-scale feature, the estimated pose $P$, and the corresponding heatmap $H$ as input for our refinement network to regress a residual $P_{\delta}$ to its ground-truth pose. 
	The final pose is obtained by $P_{\text{final}} = P + P_{\delta}$, and $L_2$ loss between the ground-truth pose and the predicted pose is applied to regress the $P_{\delta}$ for the pose refinement network:	
	\begin{equation}
		L_{\delta} = | P_{\delta} - (\Tilde{P} - P) |^2.
	\end{equation}
	
	\begin{figure*}
		\centering
		\includegraphics[width=\textwidth]{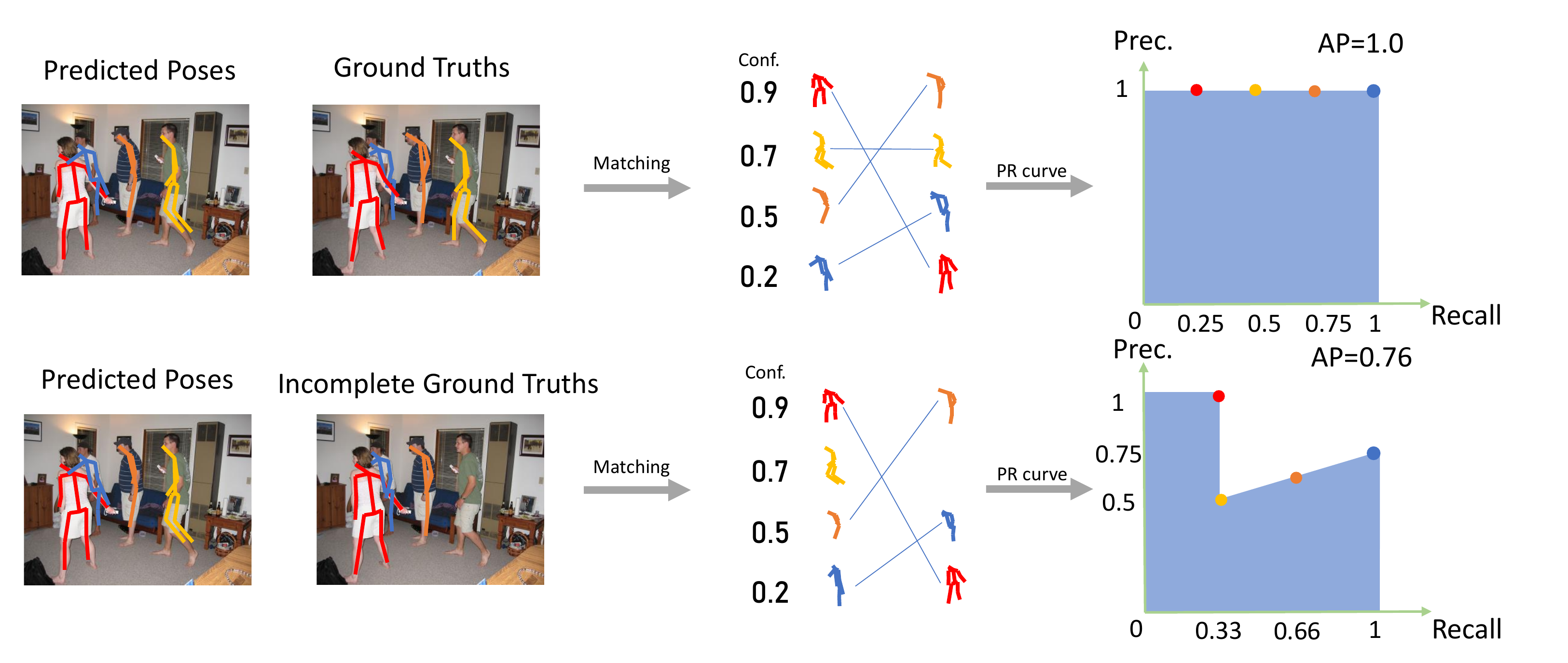}
		\caption{An example of the decreasing AP with incomplete ground truths. The first row shows the standard COCO evaluation protocol when every person is annotated in the ground truths and predicted by a pose estimation method. The resulting AP equals to 1.0 (area under the precision-recall curve). However, when one of the ground truths is missing (the yellow pose), the corresponding pose will be treated as a false positive, leading to the lower precision for the yellow pose and the subsequent ones (the yellow, orange, and blue dots in the PR curve). Therefore, the AP decreases to 0.76. Conf. and prec. stand for confidence and precision.}
		\label{fig:ap_missing_gts}
	\end{figure*}
	
	\section{Experiments}
	\label{sec:exp}
	
	\noindent \textbf{Datasets and Evaluation Metrics} We evaluate our method on COCO \cite{lin2014microsoft}, OCHuman \cite{zhang2019pose2seg}, and  CrowdPose \cite{li2019crowdpose}. Average precision (AP), AP$^{50}$, and AP$^{75}$ are used to measure overall human pose estimation accuracy on the COCO dataset. AP is used to measure cross-dataset generalization performance on the OCHuman dataset. However, AP is not an objective performance measurement if a method detects more human poses than the annotated ground truths. In fact, the predicted human poses of a method are treated as false positives if they are not annotated in the ground truths, resulting in a decrease in the AP value. 
	In particular, an example is provided in Fig.~\ref{fig:ap_missing_gts} to demonstrate the effect of the incomplete ground truths in AP calculation. In the figure, a standard COCO evaluation protocol is presented, which firstly matches the ground truths with the predicted poses and sorts the poses with their confidence, and then enumerates from the first to the end for the matched poses to calculate the precision and recall. The chart that plots the (precision, recall) pairs forms a precision-recall (PR) curve, where the AP is the area under the PR curve. If a ground-truth pose is missing, the corresponding predicted pose would be considered as a false positive, which decreases every subsequent poses' precision because of the false positive, resulting in the precision to be continuously smaller than one. As a result, the obtained AP (i.e., area under the PR curve) is lower than the one with the complete ground truths.
	Our method detects a large number of small-scale persons, which are not always annotated in the ground truths, thus resulting in a decrease in our AP values. The situation becomes even worse on the widely used COCO dataset, which contains a great number of small-scale persons without annotations as shown in Fig.~\ref{fig:qualitative_whole_img}. Therefore, simply computing AP cannot properly evaluate our method.
	
	To perform an objective evaluation, bounding box precision (BBP) is used to mitigate the negative influence of the sparse human pose annotations and act as an approximation of AP, where the human bounding box ground truths are mostly available. We also produce the corresponding bounding box recall (BBR) as an indicator of the model's ability to capture persons. 
	In particular, the evaluation of the BBP and BBR follows the standard evaluations of object detection which evaluate the AP and AR of the bounding boxes. Here we set the IoU threshold to be 0.5 which indicates the person is considered as detected if its IoU with the ground truth is larger than 0.5. In other words, the BBP and BBR are AP$^{50}$ and AR$^{50}$ in the standard evaluation of object detection. Furthermore, the evluation process of the BBP can be understood as replacing the human poses with the bounding boxes as shown in the first row of Fig.\ref{fig:ap_missing_gts} where the matching is decided by the IoU between two bounding boxes in this case.
	BBP and BBR are used as alternative measures of human pose estimation performance than AP, which are useful when human pose ground truths are missing but human bounding box annotations are available, especially for the small-scale persons, where the ground-truth annotations of the human keypoints are mostly unavailable on the COCO dataset as shown in Fig.~\ref{fig:qualitative_whole_img}.
	
	To evaluate the BBP and BBR performance on persons of different scales, COCO validation set is split into small, medium, and large-scale subsets. In particular, a person with bounding box annotation is categorized into one of the three sets based on the bounding box area, where $(0,64^2]$, $(64^2, 128^2]$, $(128^2, +\infty)$ correspond to small, medium and large, respectively. 
	
	Other than splitting COCO and CrowdPose into different subsets for BBP and BBR evaluations of every scale person, we also follow the standard protocol to split COCO test set based on the scale of the persons, where AP$^M$ and AP$^L$ represent the AP values on the subsets of medium-scale and large-scale persons respectively \cite{lin2014microsoft}.
	In addition, we provide the number of parameters and GFLOPs to measure the model size. 
	
	\begin{table*}[t]
		\footnotesize
		\centering
		\resizebox{\textwidth}{!}{
			\begin{tabular}{c|c|cccc|cccc}
				\hline
				\cline{1-10}
				\rule{0pt}{2.6ex}
				\multirow{2}{*}{\textbf{Method}} & \multirow{2}{*}{\textbf{Center type}} & \multicolumn{4}{c|}{COCO validation set} & \multicolumn{4}{c}{CrowdPose test set}\\
				\cline{3-10}
				\rule{0pt}{2.6ex}
				& & S & M & L & All & S & M & L & All\\
				\hline
				\rule{0pt}{2.6ex}
				SWAHR \cite{luo2021rethinking} (CVPR'21) & - & 25.0 & 73.0 & 82.2 & 44.5 & 38.7* & 78.5* & \underline{92.2*} &  \underline{71.2*}\\
				CGNet \cite{braso2021center} (ICCV'21) & single \& bounding box & 25.3 & \underline{77.6} & \textbf{86.4} & \underline{46.3} & 34.8* & \underline{79.6*} & 90.5* & 69.4* \\
				DEKR \cite{DEKR2021} (CVPR'21) & single \& bounding box  & 25.0 & 71.9 & 78.3 & 43.5 & \underline{41.5} & 78.0 & 87.6 & 69.9\\
				PINet \cite{wang2021robust} (NeurIPS'21) & multiple \& bounding box & \underline{27.0} & 72.6 & 80.8  & 45.4 & 38.2 & 78.0 & 89.6 & 69.7\\
				\hline
				Ours & dual \& anatomical & \textbf{35.2} & \textbf{80.3} & \underline{85.6} &  \textbf{52.9} & \textbf{58.4} & \textbf{93.2} & \textbf{96.4} &  \textbf{82.9} \\
				\hline
		\end{tabular}}
		\vspace{0.2em}
		\caption{BBR evaluations on COCO validation set and CrowdPose test set. S, M, L indicate the BBR score for small, medium, large-scale subsets. All means the BBR score for all scales (whole COCO validation set or CrowdPose test set). HRNet-W32 backbone used in all methods for coco validation set. * indicates that the HRNet-W48 backbone is used instead of the HRNet-W32 backbone. Best in \textbf{bold}, second best \underline{underlined}.}
	\label{tab:bbox_recall_all}
\end{table*}

\begin{table*}[t]
	\footnotesize
	\centering
	\resizebox{\textwidth}{!}{
		\begin{tabular}{c|c|cccc|cccc}
			\hline
			\cline{1-10}
			\rule{0pt}{2.6ex}
			\multirow{2}{*}{\textbf{Method}} & \multirow{2}{*}{\textbf{Center type}} & \multicolumn{4}{c|}{COCO validation set} & \multicolumn{4}{c}{CrowdPose test set}\\
			\cline{3-10}
			\rule{0pt}{2.6ex}
			& & S & M & L & All & S & M & L & All\\
			\hline
			\rule{0pt}{2.6ex}
			SWAHR \cite{luo2021rethinking} (CVPR'21) & - & \underline{22.1} & 64.6 & \underline{70.3}  & \underline{38.7} & \underline{33.6*} & 67.7* & 81.9* &  \underline{62.1*}\\
			CGNet \cite{braso2021center} (ICCV'21) & single \& bounding box & 21.3 & \underline{64.9} & 67.9  & 37.4 & 30.0* & \underline{68.4*} & \underline{82.4*} & 61.6* \\
			DEKR \cite{DEKR2021} (CVPR'21) & single \& bounding box  & 21.0 & 63.8 & 68.4  & 37.6 & 32.7 & 66.6 & 79.7 & 61.2\\
			PINet \cite{wang2021robust} (NeurIPS'21) & multiple \& bounding box & 20.9 & 65.0 & 15.9 & 36.1 & 30.8 & 66.0 & 80.8 & 60.7\\
			\hline
			Ours & dual \& anatomical & \textbf{30.6} & \textbf{73.2} & \textbf{77.5} &  \textbf{46.2} & \textbf{44.5} & \textbf{77.1} & \textbf{87.2} & \textbf{70.1}\\
			\hline
	\end{tabular}}
	\vspace{0.2em}
	\caption{BBP evaluations on COCO validation set and CrowdPose test set. S, M, L indicate the BBP score for small, medium, large-scale subsets. All means the BBP score for all scales (whole COCO validation set or CrowdPose test set). HRNet-W32 backbone used in all methods for coco validation set. 
		* indicates that the HRNet-W48 backbone is used instead of the HRNet-W32 backbone. Best in \textbf{bold}, second best \underline{underlined}.}
\label{tab:bbox_precision_all}
\end{table*}

\noindent \textbf{Implementation of the Multi-scale Training.}
For each image in the training, we interpolate it into different scales (256, 512, 1024). We then sequentially input the different scales of images into our model. There will be multiple forward passes with different scales of images in the training stage. This improves a model's ability to handle different scales of the persons. In the end, we gather the losses from the different scales and update the backbone.

\subsection{Quantitative Comparisons}

\noindent \textbf{Capturing Small-Scale Persons} 
Our method detects more small-scale persons, which are not annotated in the ground truths, thus resulting in a decrease in AP values, which cannot objectively reflect the performance because predicted poses without ground truths are regarded as false positives in the evaluation. Thus, the BBR evaluation is performed on the COCO and CrowdPose datasets to validate the performance of different methods for capturing small-scale persons. The results are shown in Table~\ref{tab:bbox_recall_all}, where our method outperforms all the SOTA methods in small, medium, and all scales.

\begin{table*}[h]
\footnotesize
\centering
\resizebox{\textwidth}{!}{\begin{tabular}{c|c|c|c|c|c|c|c|c|c}
	\rule{0pt}{2.6ex}
	\textbf{Group} & \textbf{Method} & Input & \#param. & GFLOPs & AP & AP$^{50}$ & AP$^{75}$ & AP$^M$ & AP$^L$\\
	\cline{1-10}
	\cline{1-10}
	\multirow{2}{*}{Top-down}
	& HRNet-W32 \cite{sun2019deep} & 384$\times$288 & - & - & 74.9 & 92.5 & 82.8 & 71.3 & 80.9\\
	& HRNet-W48 \cite{sun2019deep} & 384$\times$288 & - & - & {75.5} & {92.5} & {83.3} & {71.9} & {81.5}\\
	\cline{1-10}
	\multirow{2}{*}{One-stage}
	& FCPose (ResNet-101) \cite{mao2021fcpose} & 800 & - & - & 65.6 & 87.9 & 72.6 & 62.1 & 72.3\\
	& SPM \cite{nie2019single} & - & - & - & 66.9 & 88.5 & 72.9 & 62.6 & 73.1\\
	\cline{1-10}
	& \multicolumn{9}{c}{\rule{0pt}{2.0ex} Single-scale testing} \\
	\cline{2-10}
	\multirow{15}{*}{Bottom-up}
	& OpenPose \cite{cao2017realtime} & - & - & - & 61.8 & 84.9 & 67.5 & 57.1 & 68.2\\
	& CenterNet-HG \cite{zhou2019objects} & 512 & - & - & 63.0 & 86.8 & 69.6 & 58.9 & 70.4\\
	& PifPaf \cite{Kreiss_2019_CVPR} & - & - & - & 66.7 & - & - & 62.4 & 72.9\\
	& PersonLab \cite{papandreou2018personlab} & 1401 & - & - & 66.5 & 88.0 & 72.6 & 62.4 & 72.3\\
	& HigherHRNet-W32 \cite{cheng2020higherhrnet} & 512 & 28.5 & 47.9 & 66.4 & 87.5 & 72.8 & 61.2 & 74.2\\
	& SWAHR (HRNet-W32) \cite{luo2021rethinking} & 512 & 28.6 & 48.0 & \underline{67.9} & \textbf{88.9} & \underline{74.5} & \underline{62.4} & 75.5\\
	& DEKR (HRNet-W32) \cite{DEKR2021} & 512 & 29.6 & 45.4 & 67.3 & 87.9 & 74.1 & 61.5 & \textbf{76.1}\\
	& CGNet (HRNet-W32) \cite{braso2021center} & 512 & - & - & 67.6 & \underline{88.6} & 73.6 & 62.0 & 75.6 \\
	& PINet (HRNet-W32) \cite{wang2021robust} & 512 & - & - & 66.7 & - & - & - & -\\
	& \textbf{Ours} (HRNet-W32) & 512 & 43.5 & 46.5 & \textbf{68.5} & 87.8 & \textbf{75.2} & \textbf{63.4} & \underline{76.0}\\
	
	\cline{3-10} \rule{0pt}{2.0ex} 
	& HigherHRNet-W48 \cite{cheng2020higherhrnet} & 640 & 63.8 & 154.3 & 68.4 & 88.2 & 75.1 & 64.4 & 74.2\\
	& SWAHR (HRNet-W48) \cite{luo2021rethinking} & 640 & 63.8 & 154.6 & \underline{70.2} & \textbf{89.9} & 76.9 & 65.2 & \underline{77.0}\\
	& DEKR (HRNet-W48) \cite{DEKR2021} & 640 & 65.7 & 141.5 & 70.0 & 89.4 & \underline{77.3} & \underline{65.7} & 76.9\\
	& CGNet (HRNet-W48) \cite{braso2021center} & 640 & - & - & 69.5 & \underline{89.7} & 76.0 & 65.0 & 76.2 \\
	& \textbf{Ours} (HRNet-W48) & 640 & 79.8 & 142.7 & \textbf{71.0} & 89.5 & \textbf{78.0} & \textbf{66.1} & \textbf{78.1}\\
	\cline{2-10}
	
	& \multicolumn{9}{c}{\rule{0pt}{2.0ex} Multi-scale testing} \\
	\cline{2-10}
				\multirow{6}{*}{Bottom-up}
	& DEKR (HRNet-W32) \cite{DEKR2021} & 512 & 29.6 & 45.4 & 69.8 & \underline{89.0} & 76.6 & \underline{65.2} & 76.5\\
	& CGNet (HRNet-W32) \cite{braso2021center} & 512 & - & - & \textbf{70.3} & \textbf{90.0} & \underline{76.9} & \textbf{65.4} & \textbf{77.5} \\
	& \textbf{Ours} (HRNet-W32) & 512 & 43.5 & 46.5 & \underline{70.2} & 88.8 & \textbf{77.0} & \textbf{65.4} & \underline{77.2}\\
	
	\cline{3-10} \rule{0pt}{2.0ex} 
	& HigherHRNet-W48 \cite{cheng2020higherhrnet} & 640 & 63.8 & 154.3 & 70.5 & 89.3 & 77.2 & 66.6 & 75.8\\
	& SWAHR (HRNet-W48) \cite{luo2021rethinking} & 640 & 63.8 & 154.6 & \textbf{72.0} & \textbf{90.7} & \textbf{78.8} & \textbf{67.8} & \underline{77.7}\\
	& DEKR (HRNet-W48) \cite{DEKR2021} & 640 & 65.7 & 141.5 & 71.0 & 89.2 & 78.0 & 67.1 & 76.9\\
	& CGNet (HRNet-W48) \cite{braso2021center} & 640 & - & - & 71.4 & \underline{90.5} & 78.1 & \underline{67.2} & 77.5 \\
	& \textbf{Ours} (HRNet-W48) & 640 & 79.8 & 142.7 & \underline{71.5} & 89.1 & \underline{78.5} & \underline{67.2} & \textbf{78.1}\\
	\cline{1-10}
\end{tabular}}
\vspace{0.5em}
\caption{Evaluation on COCO test-dev set. Best in \textbf{bold}, second best \underline{underlined} for bottom-up methods. '-' indicates no data available.}
\label{tab:coco_test_dev}
\end{table*}

\begin{table*}[t]
\footnotesize
\centering
\resizebox{\textwidth}{!}{\begin{tabular}{c|c|c|c|c|c|c|c|c|c}
	\rule{0pt}{2.6ex}
	\textbf{Group} & \textbf{Method} & Input & \#param. & GFLOPs & AP & AP$^{50}$ & AP$^{75}$ & AP$^M$ & AP$^L$\\
	\cline{1-10}
	\cline{1-10}
	\multirow{2}{*}{Top-down}
	& HRNet-W32 \cite{sun2019deep} & 256$\times$192 & - & - & 74.4 & 90.5 & 81.9 & 70.8 & 81.0\\
	& HRNet-W32 \cite{sun2019deep} & 384$\times$288 & - & - & 75.8 & 90.6 & 82.5 & 72.0 & 82.7\\
	\cline{1-10}
	& \multicolumn{9}{c}{\rule{0pt}{2.0ex} Single-scale testing} \\
	\cline{2-10}
	\multirow{15}{*}{Bottom-up}
	& CenterNet-HG \cite{zhou2019objects} & 512 & - & - & 64.0 & - & - & - & -\\
	& PifPaf \cite{Kreiss_2019_CVPR} & - & - & - & 67.4 & - & - & - & -\\
	& PersonLab \cite{papandreou2018personlab} & 1401 & - & - & 66.5 & 86.2 & 71.9 & 62.3 & 73.2\\
	& HigherHRNet-W32 \cite{luo2021rethinking} & 512 & 28.5 & 47.9 & 67.1 & 86.2 & 73.0 & - & -\\
	& SWAHR (HRNet-W32) \cite{luo2021rethinking} & 512 & 28.6 & 48.0 & 68.9 & \textbf{87.8} & \underline{74.9} & \underline{63.0} & \underline{77.4}\\
	& DEKR (HRNet-W32) \cite{DEKR2021} & 512 & 29.6 & 45.4 & 68.0 & 86.7 & 74.5 & 62.1 & \textbf{77.7}\\
	& PINet (HRNet-W32) \cite{wang2021robust} & 512 & - & - & 67.4 & 86.8 & 74.0 & 62.5 & 76.3 \\ 
	& CGNet (HRNet-W32) \cite{braso2021center} & 512 & - & - & \underline{69.0} & \underline{87.7} & 74.4 & 59.9 & 75.3 \\
	& \textbf{Ours} (HRNet-W32) & 512 & 43.5 & 46.5 & \textbf{69.2} & 86.4 & \textbf{75.1} & \textbf{63.8} & 77.2\\
	\cline{3-10} \rule{0pt}{2.0ex} 
	
	& HigherHRNet-W48 \cite{luo2021rethinking} & 640 & 63.8 & 154.3 & 69.9 & 87.2 & 76.1 & - & -\\
	& SWAHR (HRNet-W48) \cite{luo2021rethinking} & 640 & 63.8 & 154.6 & 70.8 & \underline{88.5} & 76.8 & 66.3 & 77.4\\
	& DEKR (HRNet-W48) \cite{DEKR2021} & 640 & 65.7 & 141.5 & \underline{71.0} & 88.3 & \underline{77.4} & \underline{66.7} & \underline{78.5}\\
	& CGNet (HRNet-W48) \cite{braso2021center} & 640 & - & - & \underline{71.0} & \textbf{88.7} & 76.5 & 63.1 & 75.2 \\
	& \textbf{Ours} (HRNet-W48) & 640 & 79.8 & 142.7 & \textbf{72.1} & 88.3 & \textbf{78.2} & \textbf{66.9} & \textbf{79.6}\\
	\cline{2-10}
	
	& \multicolumn{9}{c}{\rule{0pt}{2.0ex} Multi-scale testing} \\
	\cline{2-10}
				\multirow{6}{*}{Bottom-up}
	& HigherHRNet-W32 \cite{luo2021rethinking} & 512 & 28.5 & 47.9 & 69.9 & 87.1 & 76.0 & - & -\\
	& SWAHR (HRNet-W32) \cite{luo2021rethinking} & 512 & 28.6 & 48.0 & \underline{71.4} & \underline{88.9} & \underline{77.8} & \underline{66.3} & \textbf{78.9}\\
	& DEKR (HRNet-W32) \cite{DEKR2021} & 512 & 29.6 & 45.4 & 70.7 & 87.7 & 77.1 & 66.2 & 77.8\\
	& CGNet (HRNet-W32) \cite{braso2021center} & 512 & - & - & \textbf{71.9} & 89.0 & \textbf{78.0} & 63.7 & 77.4 \\
	& \textbf{Ours} (HRNet-W32) & 512 & 43.5 & 46.5 & \underline{71.4} & \underline{87.6} & 77.5 & \textbf{66.6} & \underline{78.7}\\
	\cline{3-10} \rule{0pt}{2.0ex} 
	
	& HigherHRNet-W48 \cite{luo2021rethinking} & 640 & 63.8 & 154.3 & 72.1 & \underline{88.4} & 78.2 & - & -\\
	& SWAHR (HRNet-W48) \cite{luo2021rethinking} & 640 & 63.8 & 154.6 & \underline{73.2} & \textbf{89.8} & \underline{79.1} & \textbf{69.1} & \underline{79.3}\\
	& CGNet (HRNet-W48) \cite{braso2021center} & 640 & - & - & \textbf{73.3} & 89.7 & \textbf{79.2} & 66.4 & 76.7 \\
	& DEKR (HRNet-W48) \cite{DEKR2021} & 640 & 65.7 & 141.5 & 72.3 & 88.3 & \underline{78.6} & \underline{68.6} & 78.6\\
	& \textbf{Ours} (HRNet-W48) & 640 & 79.8 & 142.7 & 73.0 & 88.3 & \underline{79.1} & \underline{68.6} & \textbf{79.8}\\
	\cline{1-10}
\end{tabular}}
\vspace{0.5em}
\caption{Evaluation on COCO validation set. Best in \textbf{bold}, second best \underline{underlined} for the bottom-up methods. '-' indicates no data available.
}
\label{tab:coco_val}
\end{table*}

In particular, columns 3-6 in Table~\ref{tab:bbox_recall_all} show the BBR values of the SOTA methods and ours on different subsets of the COCO validation set.
It is observed that our method achieves a 39.1\% improvement (9.9 BBR) on the challenging subset of small-scale persons compared with the SOTA method, CGNet. 
Moreover, compared with the SOTA regression-based method PINet, we obtain a 30.3\% improvement (8.2 BBR) on the COCO small-scale subset.
Apart from our superior performance for small-scale persons, clear improvements over the SOTA methods for medium-scale persons and competitive results for large-scale persons are observed as well.
All scales combined, our improvement on the COCO validation set is 14.2\% (6.6 BBR) against the SOTA, CGNet.
The overall increase of BBR is closer to that of the small-scale subset because the COCO dataset contains more small-scale persons, where the ratio of small to medium to large-scale persons is 51:21:29.

Similarly, columns 7-10 in Table~\ref{tab:bbox_recall_all} show the BBR values on the different subsets of the CrowdPose test set.
Again, it is observed that our method demonstrates a large improvement, 50.9\% (19.7 BBR), on the challenging subset of small-scale persons compared with the SOTA, SWAHR. 
Furthermore, a 18.7\% (14.7 BBR) and a 4.5\% (4.2 BBR) improvements for medium-scale and large-scale subset are achieved by our method compared with the SOTA, which indicates our method consistently outperforms the SOTA method at each scale. 
All scales combined, our improvement over the SOTA on the CrowdPose test set is 16.4\% (11.7 BBR). 
Our improvement on the whole CrowdPose test set includes not only the small-scale persons but also medium and large-scale persons, where the small-scale persons only account for 24\% of the whole set. In particular, the ratio of small to medium to large-scale persons in the CrowdPose test set is $24:21:55$.

\noindent \textbf{Anatomical-Center Evaluation}  
We continue to compare the BBP on the COCO and CrowdPose datasets with SOTA methods. BBP exploits exactly the same dataset splits as BBR, while evaluating the precision of the captured human objects, which is an alternative measure to AP when the human pose annotations are largely missing. Similar results are shown in Table~\ref{tab:bbox_precision_all} 
, which suggests our anatomical centers outperform the SOTA methods of different center types in every scale of the persons. Column 3-6 in Table~\ref{tab:bbox_precision_all} shows the BBP values on the COCO validation set and column 7-10 in Table~\ref{tab:bbox_precision_all} shows the BBP values on the CrowdPose test set. 

To validate the proposed anatomical centers, we further focus on comparing our method with the DEKR \cite{DEKR2021} and PINet \cite{wang2021robust}, which are the SOTA regression-based methods that exploit different types of bounding box centers to identify human objects. Compared to the single-bounding-box-center method, DEKR, our anatomical centers improve the 22\% (8.6 BBP) and 14.5\% (8.9 BBP) on the COCO and Crowdpose datasets in all scales. Compared to the multiple-bounding-box-centers method, PINet, our method improves the 27.9\% (10.1 BBP) and 15.4\% (9.4 BBP) respectively in all scales. We also achieve the best BBP (and also BBR) in small, medium, and large-scale persons when compared with these two methods. In particular, PINet uses three bounding box centers, while our method simply predicts two anatomical centers which intuitively tends to decrease our performance. However, even in this circumstance, we outperform DEKR and PINet at each scale, which clearly demonstrates that our proposed dual anatomical centers are more effective than the single or multiple bounding box centers. A similar result is observed in the AP evaluation in the next section as well. 

Apart from DEKR and PINet, we also compared with SWAHR, which is the best-performed existing method in Table~\ref{tab:bbox_precision_all}. 
It is observed that our method achieves a significant increase of BBP for small-scale persons compared with SWAHR, 38.4\% (8.5 BBP) on the COCO validation set and 32.4\% (10.9 BBP) on the CrowdPose test set.
It is worth noting that the reported numbers of CGNet and SWAHR on the CrowdPose dataset are based on their models with HRNet-W48 backbone, where our results are based on a smaller backbone, HRNet-W32, to be comparable to DEKR and PINet. Even with such a disadvantage, we still outperform both CGNet and SWAHR by a large margin.

\noindent \textbf{Overall Performance}  
Apart from the bounding box recall and precision evaluations at different scales of persons, we also provide the AP comparisons of the overall performance evaluation on the COCO validation and test-dev set as shown in  Table~\ref{tab:coco_test_dev} and Table~\ref{tab:coco_val}. Note that the AP values cannot objectively reflect our performance because our method detects a large number of small-scale that are not annotated, which are counted as false positives in AP evaluation. However, even in this circumstance, benefiting from the proposed \multi{multi-scale training}, our method outperforms existing bottom-up methods consistently in single-scale testing on both COCO validation set and test set. In particular, we achieve 1.1 improvement of AP with HRNet-W48 backbone in Table~\ref{tab:coco_val}, resulting in 1.5\% increase against the SOTA method \cite{DEKR2021} and 0.2 improvement of AP compared to SOTA with HRNet-W32 backbone. In test-dev set shown in Table~\ref{tab:coco_test_dev}, our improvement is 0.6 AP over SWAHR's with HRNet-W32 backbone, which is significant by comparing with the latest SOTA method, where PINet's improvement over HigherHRNet (the previous SOTA) is 0.3 AP with HRNet-W32 backbone. The 0.6 AP increase is significant but not incremental given the fact that 2D pose estimation has been extensively studied for many years.


Specifically, we achieve 71.0 AP with HRNet-W48 backbone in Table~\ref{tab:coco_test_dev}, which is a 1.1\% increase against the SOTA method, SWAHR. The increase should be considered together with the BBR evaluation given the missing annotations in the ground truths, where SWAHR's BBR is 44.5 and ours is 52.9 on the COCO validation set, another 11.6\% improvement is achieved by our method compared with SWAHR, the SOTA method focused on small-scale persons handling.
Besides single-scale testing, we also conducted the multi-scale testing with different backbones in Table~\ref{tab:coco_test_dev} and Table~\ref{tab:coco_val}. We can also achieve comparable performance compared to SOTA methods even if our method is designed for enhancing the single-scale testing.

\begin{table*}[t]
\footnotesize
\centering
\resizebox{\textwidth}{!}{
\begin{tabular}{c|c|c|c|c|c|c|c}
	\hline
	\cline{1-8}
	\rule{0pt}{2.6ex}
	\textbf{Metric} & AE \cite{newell2017associative} & HGG \cite{jin2020differentiable} & DEKR \cite{DEKR2021} & SWAHR \cite{luo2021rethinking} & CGNet \cite{braso2021center} & PINet \cite{wang2021robust} & Ours\\
	\hline
	\rule{0pt}{2.6ex}
	AP & 29.5 & 34.8 & 36.4 & 39.5 & 38.6 & 37.2 & \textbf{40.3}\\
	\hline
\end{tabular}}
\vspace{0.2em}
\caption{Quantitative comparison on OCHuman test set. All results are based on single-scale testing with flipping tests. HRNet-W32 backbone used in all methods.}
\label{tab:ochuman}
\end{table*}

\noindent \textbf{Performance of Cross-Dataset Generalization}  
We perform a quantitative comparison on the OCHuman dataset \cite{zhang2019pose2seg}, which focuses on the crowded scenes with human interactions. Since OCHuman is a new dataset, where several SOTA methods do not report their performance and do not release their pre-trained models on it \cite{DEKR2021,luo2021rethinking}, we follow the evaluation of PINet \cite{wang2021robust} to perform testing on OCHuman test set with the models trained on the COCO dataset. Since the OCHuman training set is not used, this comparison evaluates the cross-dataset generalization capability of different methods as shown in Table~\ref{tab:ochuman}. We observe that our method outperforms all the existing methods and our superior performance shows our method has better generalization capability compared with other SOTA methods in the crowded scenes with occlusions.

\subsection{Qualitative Evaluations}

\input{figure_tex/qualitative_results}


Fig. {\ref{fig:qualitative_result}} shows the qualitative results of our method compared with DEKR {\cite{DEKR2021}} and SWAHR {\cite{luo2021rethinking}}. The first two rows show a surfing image with four small-scale persons, where DEKR predicts one person and SWAHR predicts two. In contrast, our method correctly predicts the poses for all the four persons, while only one person is annotated in the GT. The {\nth{3}} and {\nth{4}} rows show a sports scene with large to small-scale persons. In the zoom-in view, we can see that both DEKR and SWAHR predict two poses, GT has three persons annotated, and our method predicts the three persons correctly and three additional persons. The last two rows show a third example that our method can detect small-scale persons robustly and avoid missing or duplicate predictions.

\begin{figure*}
    \centering
    
    \begin{minipage}[c]{0.240\linewidth}
    \centerline{\scriptsize{DEKR}}
	\vspace{0.3em}
	\end{minipage}
	\begin{minipage}[c]{0.240\linewidth}
    \centerline{\scriptsize{SWAHR}}
	\vspace{0.3em}
	\end{minipage}
	\begin{minipage}[c]{0.240\linewidth}
    \centerline{\scriptsize{Ours}}
	\vspace{0.3em}
	\end{minipage}
	\begin{minipage}[c]{0.240\linewidth}
    \centerline{\scriptsize{Ground truth}}
	\vspace{0.3em}
	\end{minipage}
    
    \begin{minipage}[c]{0.240\linewidth}
		\includegraphics[width=\linewidth]{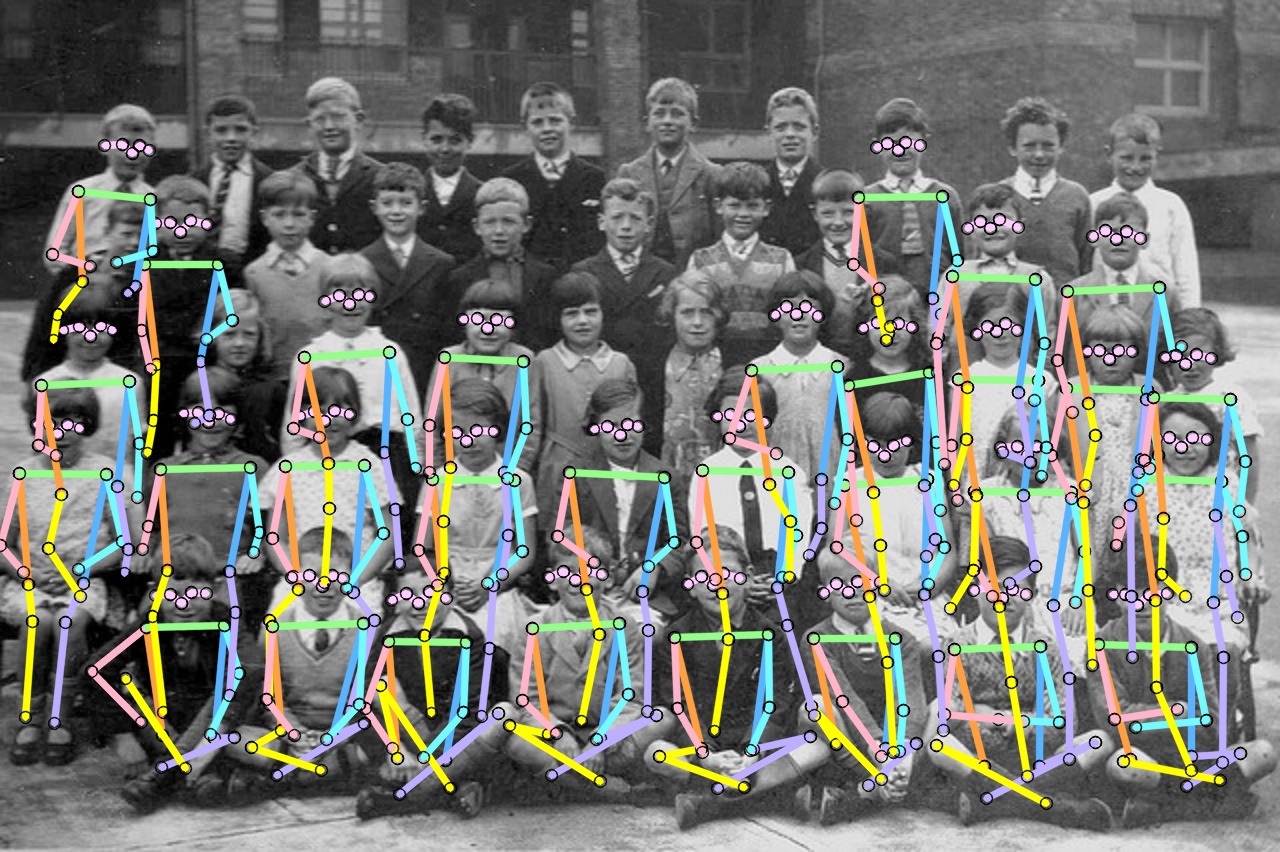}\vspace{0.1em}
	\end{minipage}
	\begin{minipage}[c]{0.240\linewidth}
		\includegraphics[width=\linewidth]{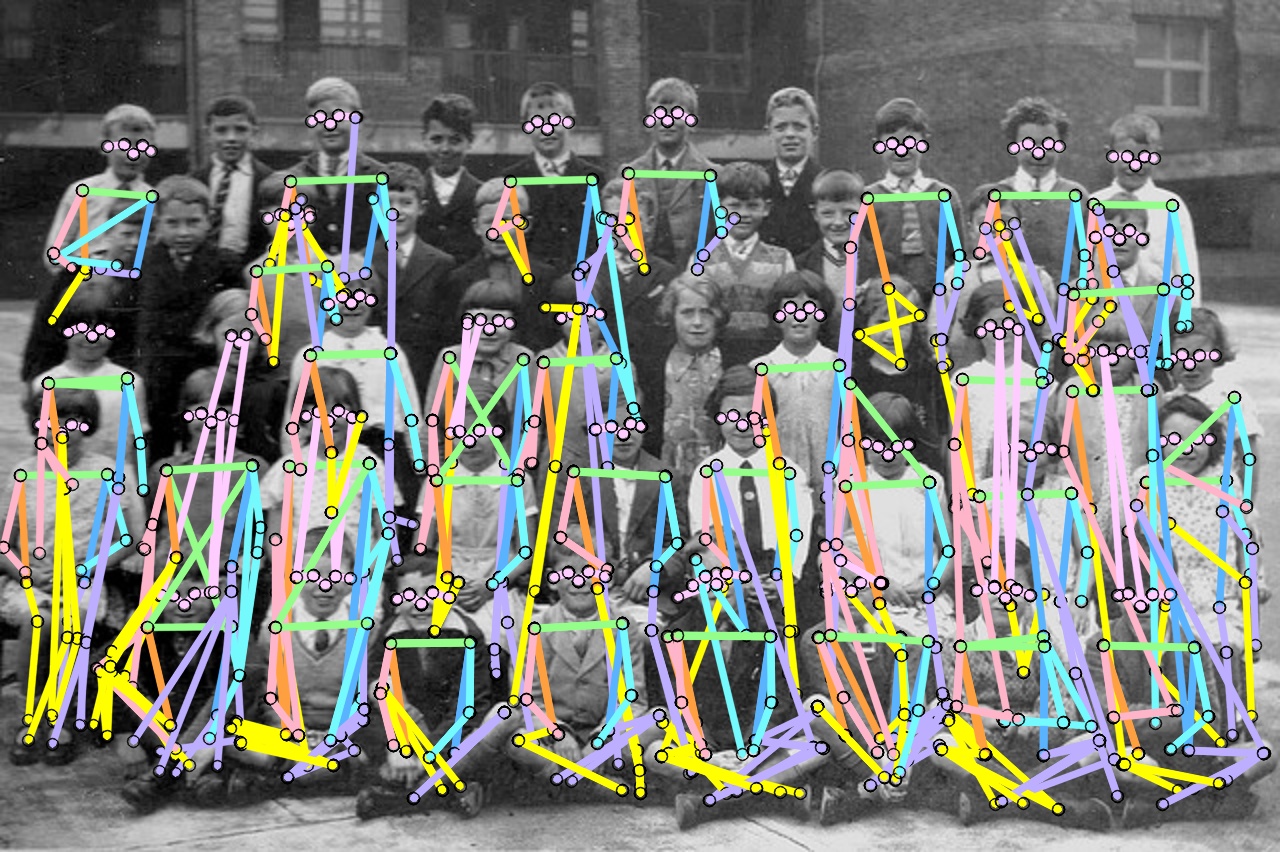}\vspace{0.1em}
	\end{minipage}
	\begin{minipage}[c]{0.240\linewidth}
		\includegraphics[width=\linewidth]{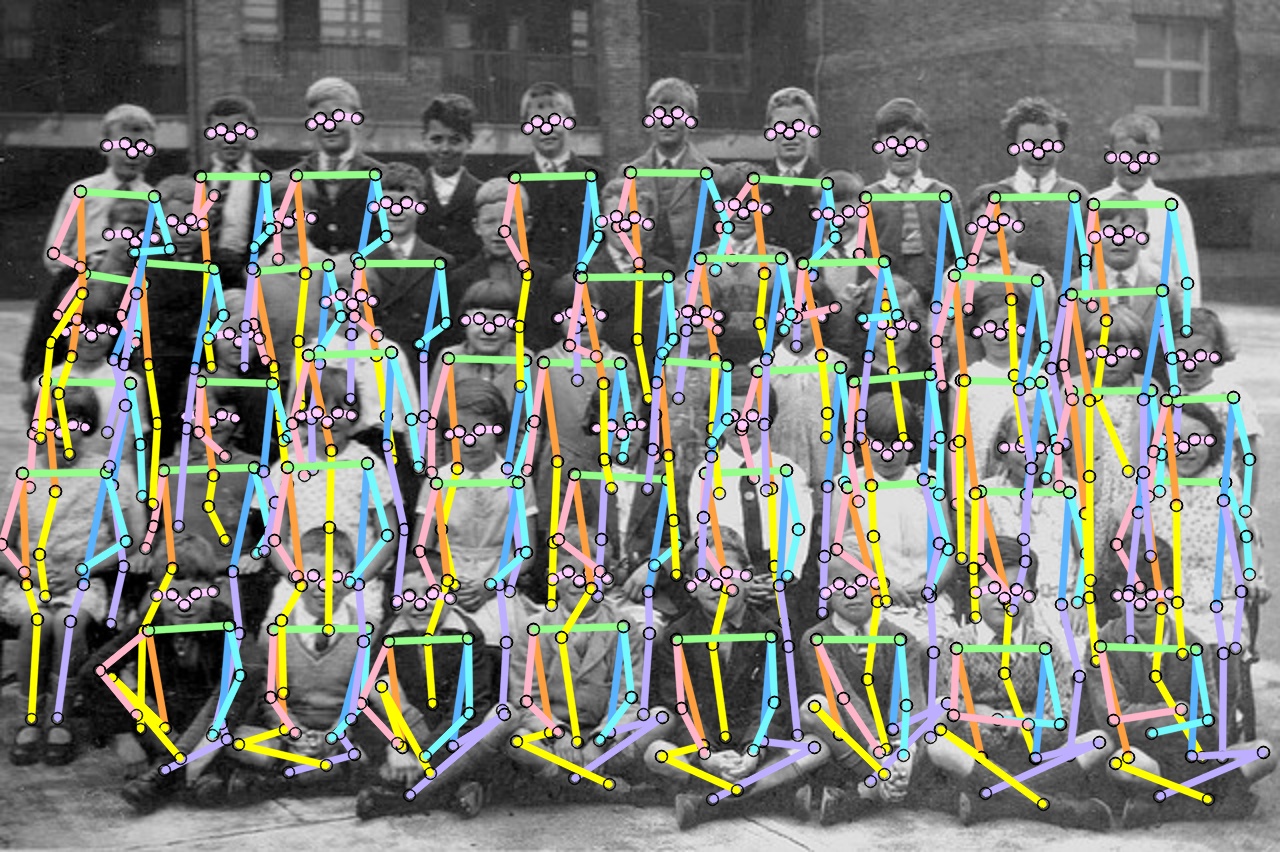}\vspace{0.1em}
	\end{minipage}
	\begin{minipage}[c]{0.240\linewidth}
		\includegraphics[width=\linewidth]{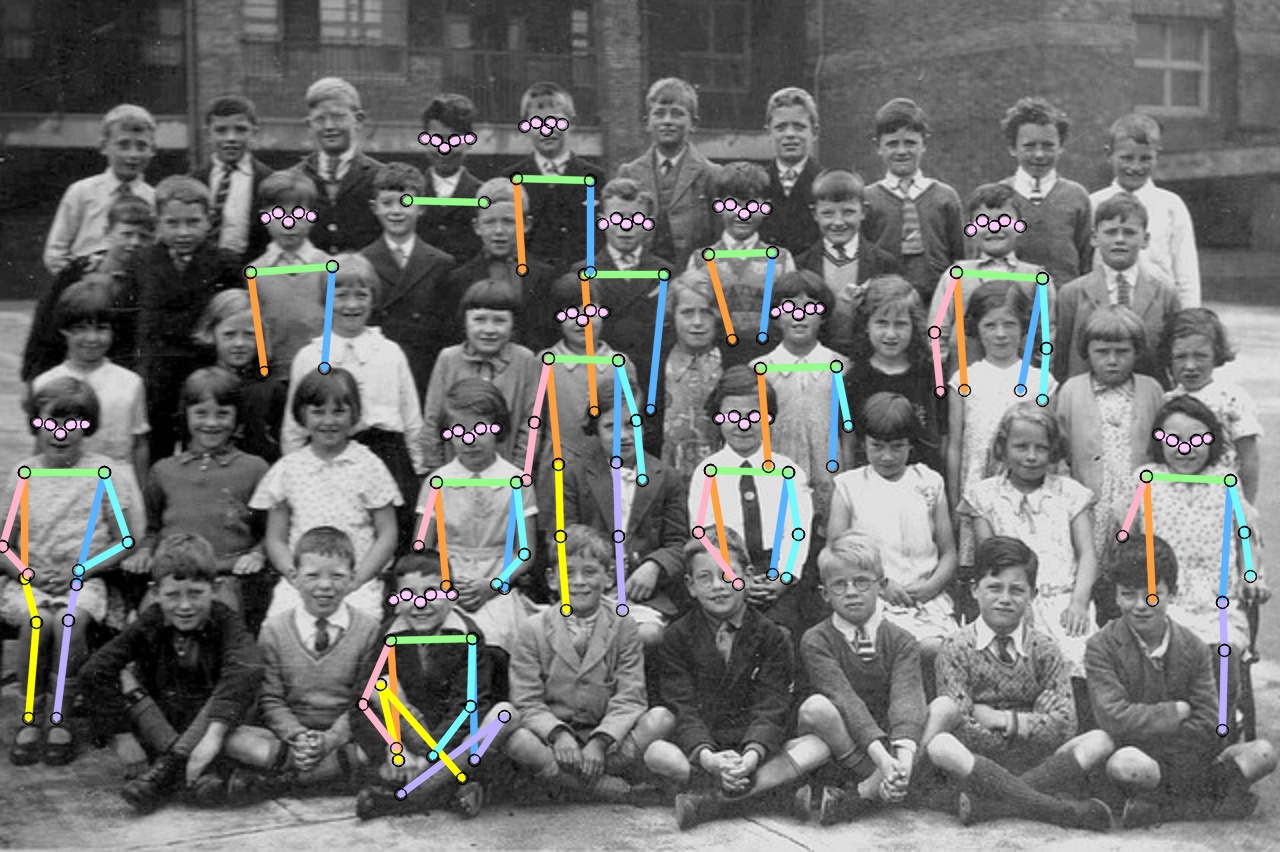}\vspace{0.1em}
	\end{minipage}
	
	\begin{minipage}[c]{0.240\linewidth}
		\includegraphics[width=\linewidth]{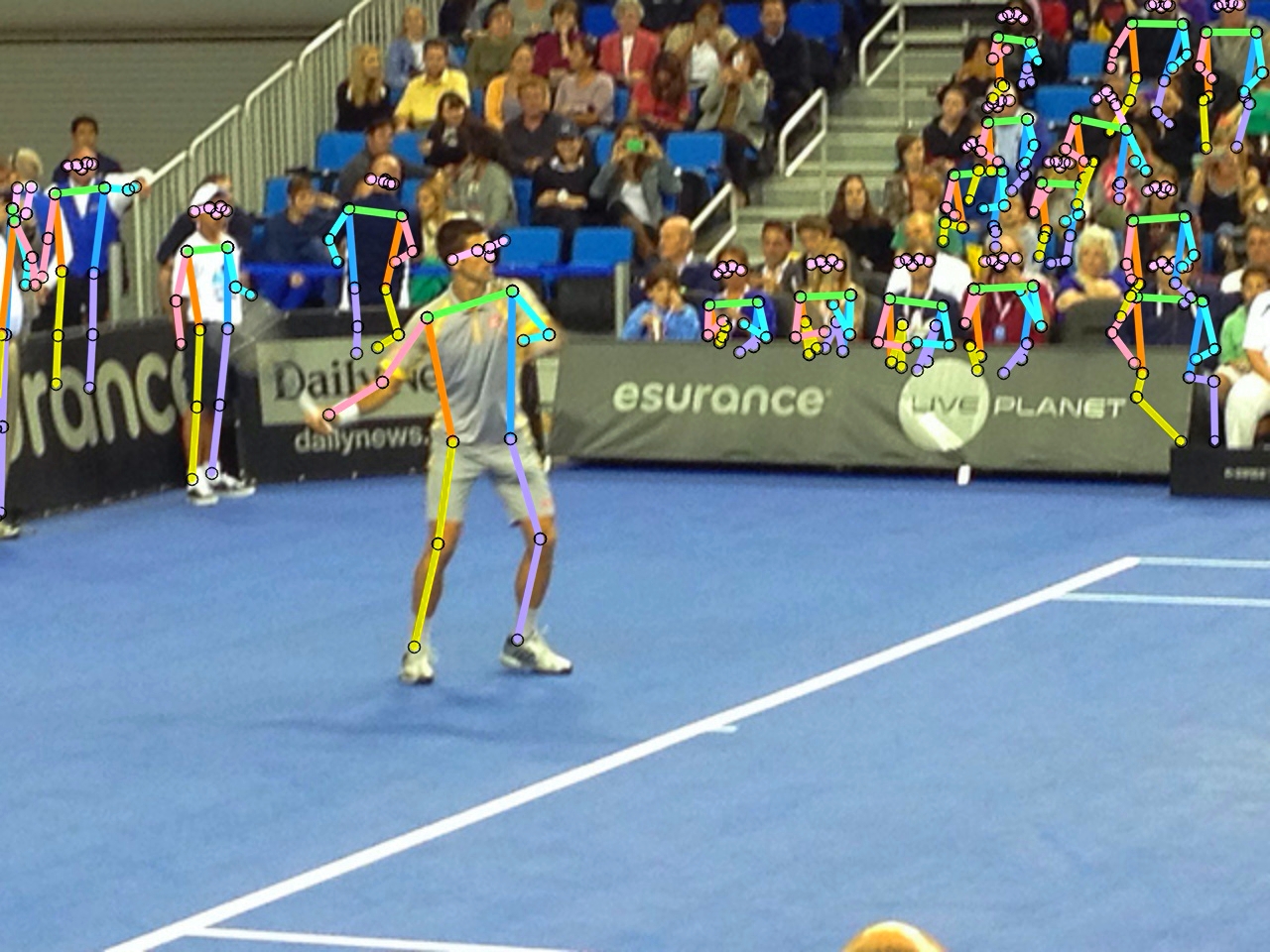}\vspace{0.1em}
	\end{minipage}
	\begin{minipage}[c]{0.240\linewidth}
		\includegraphics[width=\linewidth]{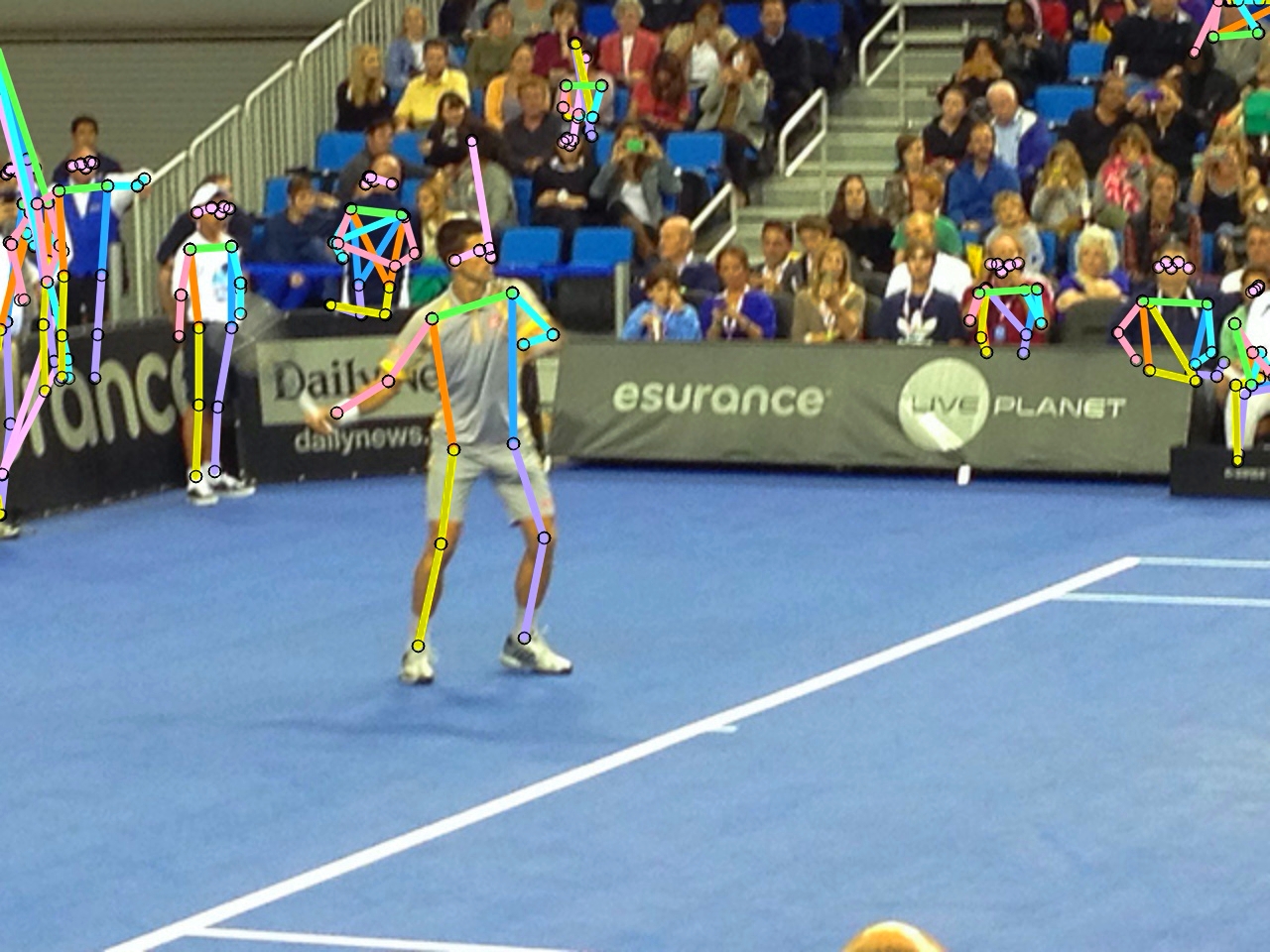}\vspace{0.1em}
	\end{minipage}
	\begin{minipage}[c]{0.240\linewidth}
		\includegraphics[width=\linewidth]{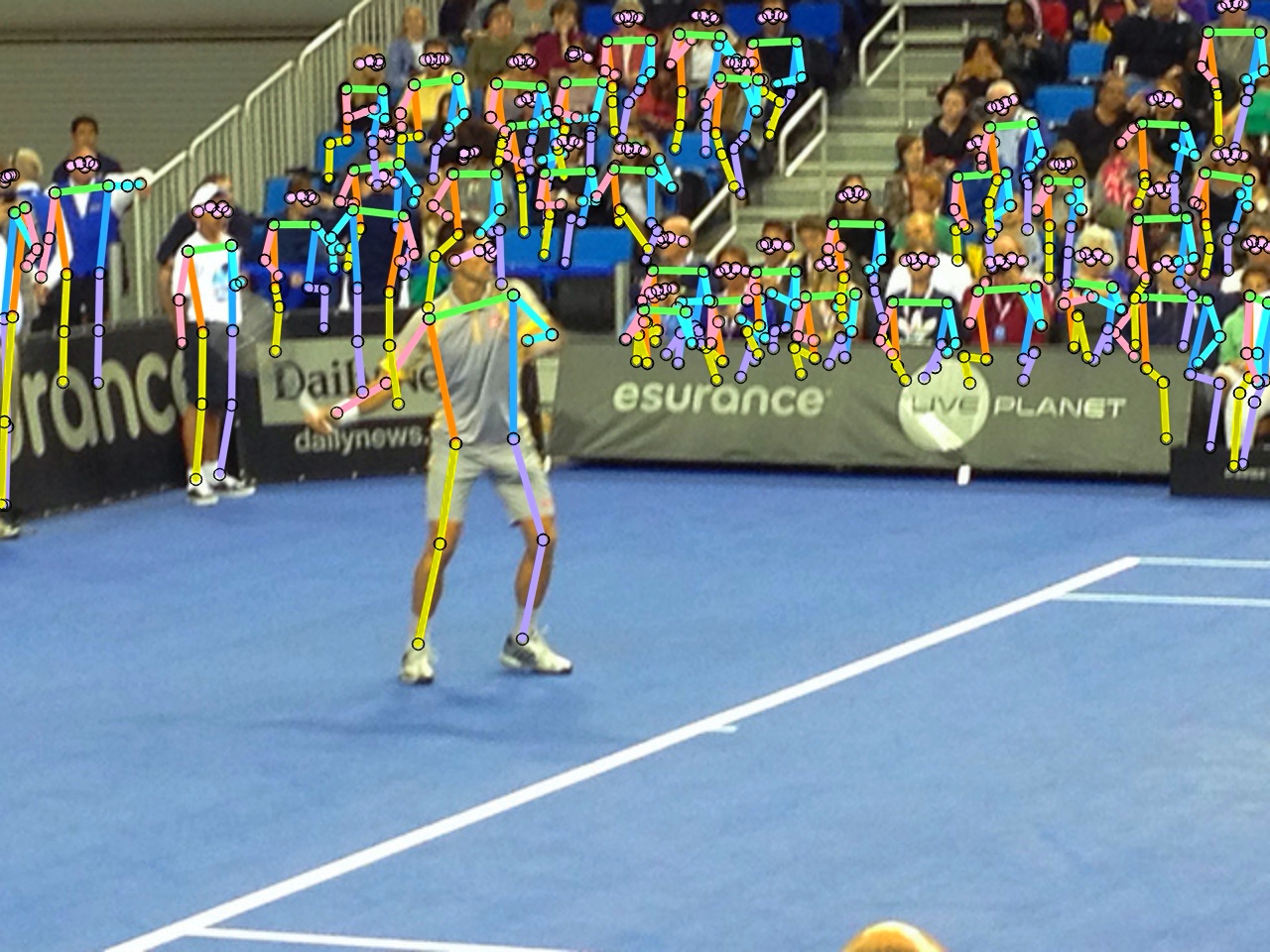}\vspace{0.1em}
	\end{minipage}
	\begin{minipage}[c]{0.240\linewidth}
		\includegraphics[width=\linewidth]{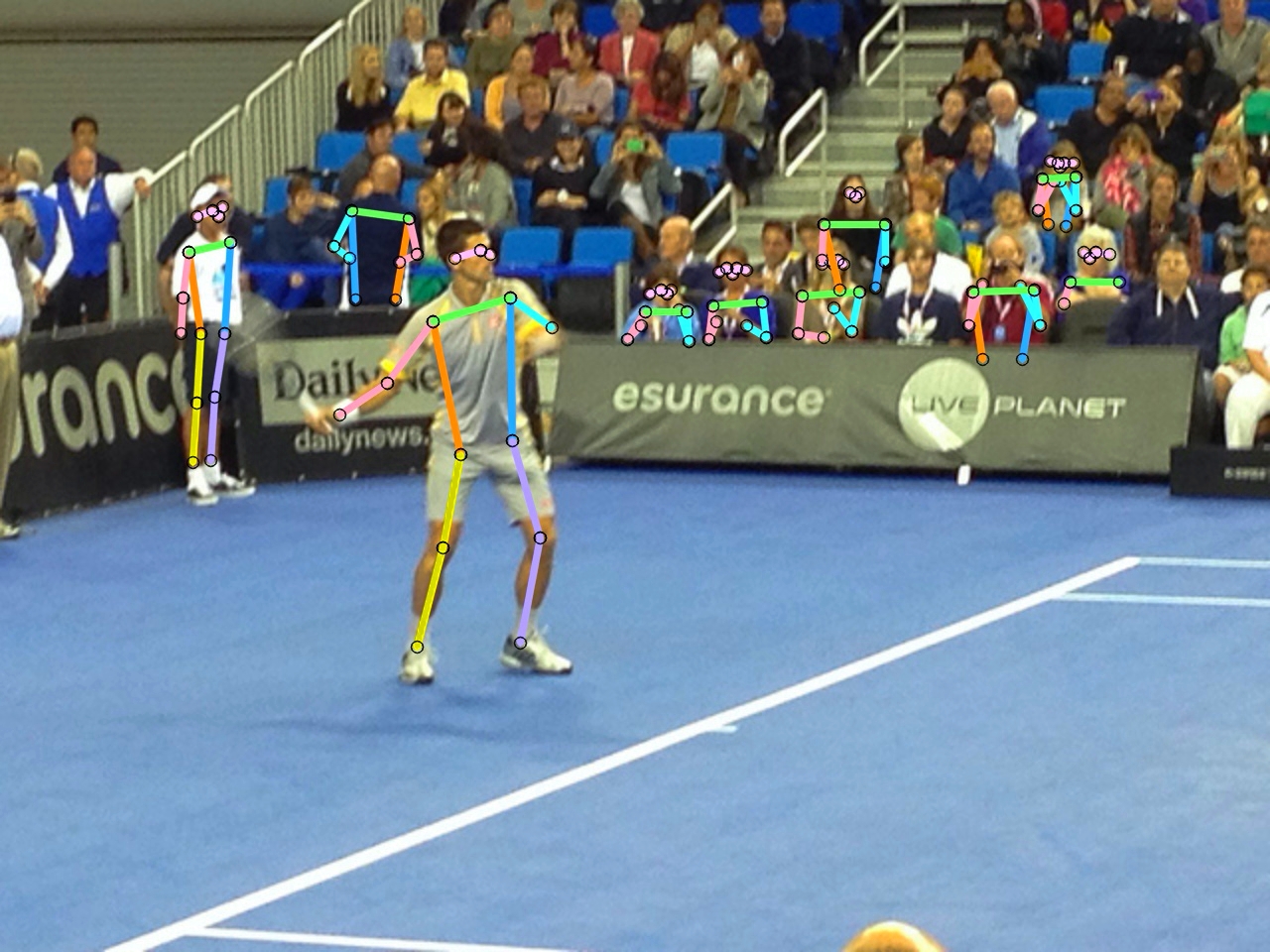}\vspace{0.1em}
	\end{minipage}
	
    \caption{Qualitative comparison on the COCO validation set with the SOTA methods: DEKR \cite{DEKR2021} and SWAHR \cite{luo2021rethinking} in crowd scenes. Each row shows the pose estimation result of each method on the whole image.}
    
    \label{fig:qualitative_whole_img}
\end{figure*}

Fig. {\ref{fig:qualitative_whole_img}} shows additional qualitative results in crowd scenes, where many persons are highly overlapped in both examples and most of the persons are small-scale in the second row. Compared to the SOTA methods, our proposed method can produce accurate pose estimation in the crowd scenes. Note that the GT annotation is sparse, where many persons that our method predicts do not have corresponding GT.

\subsection{Ablation Studies}

The effectiveness of major components of our method is validated in Table~\ref{tab:ablation}. We remove all the proposed components from our method as a baseline approach, which uses a single bounding box center for human detection with non-maximum suppression (NMS) to filter out duplicate human poses. 
The baseline is based on single-scale testing without flipping as stated in Table~\ref{tab:ablation}, where flipping is not used to remove its influence on performance (not from proposed components). For example, DEKR's AP drops from 68.0 to 62.3 when flipping is removed.

\begin{table}[t]
\footnotesize
\centering
\begin{tabular}{c|c|c|c}
\hline
\cline{1-4}
\rule{0pt}{2.6ex}
\textbf{Method} & AP & AP$^M$ & AP$^L$\\
\cline{1-4}
\rule{0pt}{2.6ex}
baseline & 62.4 & 56.2 & \underline{74.1}\\
+ two centers & 61.3 & 55.8 & 71.2\\
+ multi-scale training & 62.8 & 57.1 & 73.0\\
+ pose confidence & 63.6 & 57.4 & 73.5\\
+ pose similarity & \underline{64.0} & \underline{57.7} & 74.0\\
+ pose refinement  & \textbf{64.4} & \textbf{58.1} & \textbf{74.3} \\
\cline{1-4}
\end{tabular}
\vspace{0.3em}
\caption{Ablation study on COCO validation set. Best in \textbf{bold}, second best \underline{underlined}. Numbers based on single-scale testing without flipping.}
\label{tab:ablation}
\end{table}

On top of the baseline, we add each proposed component at a time and show the resulting performance. It is observed that most of the components help to improve the overall accuracy, where the multi-scale testing brings the largest improvement (1.5 AP). Pose confidence ranks the 2\textsuperscript{nd} in terms of AP improvement, which helps to increase AP by 0.8. Note that in the second row of Table~\ref{tab:ablation}, the result of '+ two centers' reduces the AP compared with the baseline,
since the average of two centers' confidence is used as the pose confidence to replace our proposed pose confidence for ablation study purposes. 
Once the pose confidence is added, the AP is improved and better than the baseline. 

\noindent \textbf{Validation of the Dual Centers} Fig.~\ref{fig:ablation_tables} (left) shows a comparison of the proposed dual centers against the baseline (one bounding box center) used in an existing approach \cite{DEKR2021}, where the AP and recall changes over the baseline are provided. 
The recall measures the percentage of persons detected. In particular, the statistics are computed based on OKS larger than 0.5, if a pose has an OKS value larger than 0.5 compared to a ground truth, it is counted as a detected person, thus increasing the recall value. 
The AP and recall of the baseline are 62.4 and 89.5. 
It is observed that our dual centers approach outperforms the baseline (0.8 recall), which means our approach captures 0.8\% more annotated persons that is 51 persons out of 6352 annotated in COCO val set.
In fact, our method detects many more persons that are not annotated as shown in Fig.~\ref{fig:teaser_example}.


The effect of using different numbers of the anatomical centers is explored as well in Fig.~\ref{fig:ablation_tables} (left), where we show the results of using either one of the two centers or three centers (the third one is the left shoulder). Using either one of the two centers is worse than the proposed dual-centers approach, and the result of three centers marginally improves recall but hurts the AP. Such evidence supports our choice of using dual anatomical centers. 

\noindent \textbf{Pose Similarity and Confidence}
The effectiveness of the proposed pose similarity and pose confidence is validated in Fig.~\ref{fig:ablation_tables} (right). The baseline's AP is 62.8, where the average of two centers' confidence is used as pose confidence and the poses are grouped and selected based on all joints' average Euclidean distance. On top of the baseline, we use our proposed pose similarity to group the poses, referred as Method 1. Only using the proposed pose confidence without the proposed pose similarity is Method 2, and using both is Method 3, where our pose similarity and confidence bring 1.2 AP improvement against the baseline. 

\pgfplotsset{width=6.2cm,height=4.1cm,compat=1.9}

\begin{figure*}[t]
\centering
\begin{tikzpicture}
\scriptsize
\begin{axis}[
    title={Dual centers evaluation},
    xlabel={Method},
    ylabel={Change over baseline},
    xmin=1, xmax=4,
    ymin=-3, ymax=3.4,
    xtick={1, 2, 3, 4},
    ytick={-3, -1.5, 0, 1.5, 3},
    legend pos=north west,
    ymajorgrids=true,
    grid style=dashed,
]
\addplot[
    color=red,
    mark=star,
    ]
    coordinates {
    (1,-3.0)(2,-2.3)(3,1.6)(4,1.3)
    };
    \legend{AP change}

\addplot[
    color=blue,
    mark=o,
    ]
    coordinates {
    (1,-2.7)(2,-1.6)(3,0.8)(4,0.9)
    };
    \addlegendentry{Recall change}
\end{axis}
\end{tikzpicture}
\hskip 9pt
\begin{tikzpicture}
\scriptsize
\begin{axis}[
    title={Pose similarity and confidence eval.},
    xlabel={Method},
    ylabel={AP},
    xmin=1, xmax=3,
    ymin=62.5, ymax=64.5,
    xtick={1, 2, 3},
    ytick={62.5, 63, 63.5, 64},
    legend pos=north west,
    ymajorgrids=true,
    grid style=dashed,
]
\addplot[
    color=red,
    mark=square,
    ]
    coordinates {
    (1,63.1)(2,63.6)(3,64.0)
    };
    \legend{Ours}
\addplot[
    color=blue,
    mark=,
    ]
    coordinates {
    (1,62.8)(2,62.8)(3,62.8)
    };
    \addlegendentry{Baseline}
\end{axis}
\end{tikzpicture}
\caption{Evaluations of two proposed components on COCO validation set. Left: evaluation of the dual centers approach; Method 1-4: head center only, body center only, the proposed dual centers, and three centers. Right: evaluation of the proposed pose similarity and confidence; Method 1-3: baseline + pose similarity, baseline + pose confidence, baseline + both. Same testing condition as Table~\ref{tab:ablation} in both evaluations.}
\label{fig:ablation_tables}
\end{figure*}
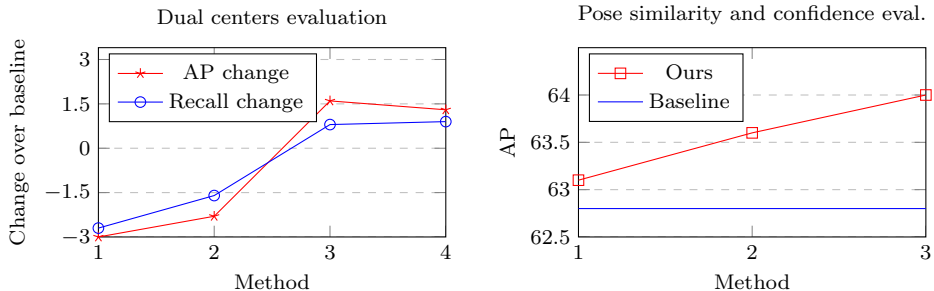

\subsection{Limitations and Failure Cases}
\begin{figure}[t]
\centering
\includegraphics[width=\linewidth]{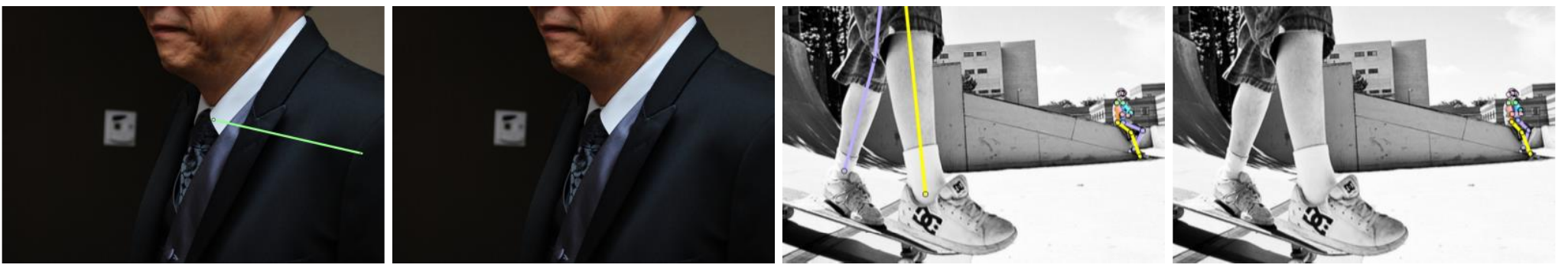}
\caption{Failure cases of our method on COCO validation set. Column 1, 3: the ground truth, column 2, 4: results of our method.}
\label{fig:failure_case}
\end{figure}

Our method may have problems processing out-of-boundary large-scale persons. If one or several large-scale persons are present and more than half of the body of them is out of the image boundary, our method could fail. The reason lies in the proposed dual-centers approach that tries to find the head and body centers for a person. A few failure cases are shown in Fig.~\ref{fig:failure_case} and we plan to solve this problem in the future.

\section{Conclusion}
\label{sec:conclusion}
We propose a novel bottom-up multi-person pose estimation method and focus on the performance for small-scale persons. 
First, we introduce \multi{multi-scale training} to improve the accuracy of small-scale persons with single-scale testing. 
Second, our dual anatomical center approach allows the human poses to be estimated accurately and reliably due to our proposed human centers being anatomically defined and consistent to visual context. 
Altogether, our method demonstrates its superior performance on the BBP, BBR, and AP metrics across different public datasets, especially in the challenging small-scale-person scenarios.


\section*{Acknowledgment}

This research is supported by the National Research Foundation, Singapore under its Strategic Capability Research Centres Funding Initiative. Any opinions, findings and conclusions or recommendations expressed in this material are those of the author(s) and do not reflect the views of National Research Foundation, Singapore.

\bibliographystyle{elsarticle-num-names.bst}
\bibliography{egbib}

\vfill

\end{document}